\documentclass[a4paper,11pt,fleqn]{article}

\usepackage{algorithm}
\usepackage{algpseudocode}
\usepackage{amsmath}
\usepackage{graphicx}
\usepackage{rotating}
\usepackage[authoryear]{natbib}
\usepackage{booktabs}
\usepackage{longtable}
\usepackage{acro}
\usepackage{cleveref}
\usepackage{caption}
\usepackage{subcaption}
\usepackage{pgfplots}
\usepackage{booktabs}
\usepackage{geometry}
\usepackage{authblk}
\usepackage{comment}

\usepackage{todonotes}

\newcommand{\new}[1]{\textcolor{black}{#1}}

\DeclareAcronym{los}{
	short=LOS,
	long=length-of-stay,
}
\DeclareAcronym{ml}{
	short=ML,
	long=machine learning,
}
\DeclareAcronym{ed}{
	short=ED,
	long=emergency department,
}
\DeclareAcronym{rmse}{
	short=RMSE,
	long=root mean square error,
}
\DeclareAcronym{rmsle}{
	short=RMSLE,
	long=root mean square logarithmic error,
}
\DeclareAcronym{mae}{
	short=MAE,
	long=mean absolute error,
}
\DeclareAcronym{mse}{
	short=MSE,
	long=mean square error,
}
\DeclareAcronym{opr}{
	short=OT,
	long=operating theatre,
}
\DeclareAcronym{or}{
	short=OR,
	long=operation research,
}
\DeclareAcronym{milp}{
	short=MILP,
	long=mixed integer linear programming,
}
\DeclareAcronym{mss}{
	short=MSS,
	long=master surgical schedule,
}
\DeclareAcronym{icu}{
	short=ICU,
	long=intensive care unit,
}
\DeclareAcronym{saa}{
	short=SAA,
	long=sample average approximation,
}
\DeclareAcronym{mdp}{
	short=MDP,
	long=Markov decision process,
}

\begin{document}

\title{Prediction accuracy versus rescheduling flexibility in elective surgery management}

\author[1]{Pieter Smet}
\author[2,3,4]{Martina Doneda}
\author[3]{Ettore Lanzarone}
\author[2]{Giuliana Carello}

\affil[1]{Department of Computer Science, KU Leuven, Gent, Belgium}
\affil[2]{Department of Electronics, Information and Bioengineering, Politecnico di Milano, Milan, Italy}
\affil[3]{Department of Management, Information and Production Engineering, University of Bergamo, Dalmine (BG), Italy}
\affil[4]{Institute for Applied Mathematics and Information Technologies, National Research Council of Italy, Milan, Italy}

\maketitle

\abstract{
\noindent The availability of downstream resources plays is critical in planning the admission of elective surgery patients. The most crucial one is inpatient beds.
To ensure bed availability, hospitals may use \ac{ml} models to predict patients' \ac{los} in the admission planning stage.
However, the real value of the\ac{los} for each patient may differ from the predicted one, potentially making the schedule infeasible.
To address such infeasibilities, it is possible to implement rescheduling strategies that take advantage of operational flexibility.
For example, planners may postpone admission dates, relocate patients to different wards, or even transfer patients who are already admitted among wards.
A straightforward assumption is that better \ac{los} predictions can help reduce the impact of rescheduling.
However, the training process of \ac{ml} models that can make such accurate predictions can be very costly.
Building on previous work that proposed simulated \ac{ml} for evaluating data-driven approaches, this paper explores the relationship between \ac{los} prediction accuracy and rescheduling flexibility across various corrective policies.
Specifically, we examine the most effective patient rescheduling strategies under \ac{los} prediction errors to prevent bed overflows while optimizing resource utilization.}


\section{Introduction}
\label{sec:losintroduction}

Growing pressure on healthcare systems, along with increasingly stringent performance standards, have led hospital administrators to pursue cost reductions by improving operational efficiency.
As a result, surgical departments -- critical to hospital operations yet expensive to run -- have become a central focus of research, particularly in the planning of activities within \acfp{opr}.
When planning the admission of patients for elective surgery, accounting for bed availability during the patients' \acf{los}\footnote{We define \ac{los} as the number of days a patient remains in an inpatient ward following a surgical procedure.} is critical, as surgery schedules optimized without considering this downstream resource may easily become infeasible in practice due to, for example, excessive demand for inpatient beds \citep{hans2023}.
Typically, \ac{los} is estimated based on the clinical reason for admission, relying on the assumption that each procedure has a standard recovery time \citep{fetter1980, shea1995, grubinger2010}.
However, clinical practice has shown that \ac{los} is influenced by various factors, such as patient demographics, unforeseen complications, hospital-acquired infections, and discharge planning policies --which, in turn, may depend on external factors such as family support arrangements \citep{van2015} and the availability of social care or community nursing support \citep{mcmullan2004}.
Such factors can either extend or shorten \ac{los}, thereby deviating from the initial estimates made at the time of admission.
Such deviations can introduce several issues: longer stays may strain bed capacity, leading to the cancellation of elective surgeries, while shorter stays may result in inefficient resource utilization and contribute to long waiting lists.
Both outcomes are costly and can negatively impact patients awaiting timely care.
To mitigate the effects of poor \ac{los} estimates, hospital managers can resort to rescheduling strategies that leverage varying degrees of operational flexibility \citep{akbarzadeh2024}.

In this paper, we propose a methodology to evaluate the impact of \ac{los} prediction accuracy in data-driven decision support systems to schedule elective patient admissions.
Specifically, we look at a predict-then-optimize approach \citep{robinson1966} in which an \acf{ml} model predicts patients' \acp{los}, which are then used as input for a deterministic optimization model that generates patient admission schedules.
While more accurate predictions generally lead to better outcomes, achieving higher accuracy often comes with exponentially increasing training costs \citep{tulabandhula2013}.
This trade-off between prediction accuracy and training cost is often difficult to assess in advance.
However, given the effort needed to collect data and train predictive models, it is worthwhile to evaluate the potential advantages of using \ac{los} estimations before committing resources to develop an \ac{ml}-based solution approach.
\citet{vancroonenburg2016study} obtained such insights for the patient-to-room assignment problem.
Through a computational study, they demonstrated how solution quality decreases as \ac{los} estimates become worse.
Although many predictive methods for \ac{los} estimation have been proposed over the years, the impact of various uncontrollable (external) factors prevents any single technique from being universally effective \citep{awad2017}.

Our work builds upon the framework introduced by \citet{doneda2024rob}, which assumes the existence of a predictive \ac{ml} model with a well-defined error metric.
Instead of training and testing actual \ac{ml} models, our approach simulates the predictions that an \ac{ml} model might generate at different levels of performance.
By controlling the error and generating simulated predictions, we evaluate how different rescheduling strategies are impacted by \ac{los} prediction accuracy.
This way, we study four rescheduling strategies that exploit different types of operational flexibility to restore feasibility when admission schedules are disrupted by inaccurate \ac{los} predictions.
Using data from the largest university hospital in Belgium, we conduct a computational study to gain insights into the interplay between \ac{los} prediction accuracy and patient admission rescheduling.

The remainder of this paper is organized as follows.
In \Cref{sec:losliterature}, we review literature on surgery scheduling considering \ac{los} and on \ac{los} prediction with \ac{ml}.
In \Cref{sec:losproblem}, we present the formal description of the scheduling problem and introduce four different rescheduling policies.
The simulated prediction framework is presented in \Cref{sec:losprediction}.
The setup of the computational experiments is described in \Cref{sec:losexpml}, while the obtained results are analyzed and discussed in \Cref{sec:losresults}.
\Cref{sec:losconclusions} concludes the paper by summarizing our main insights and identifying potential directions for follow-up research.

\section{Literature review}
\label{sec:losliterature}

\Cref{sec:schedulinglit} provides a comprehensive overview of the operations research literature on surgery scheduling, while \Cref{sec:lospredictionlit} reviews the approaches that have been proposed to predict patient \ac{los} in various settings.

\subsection{Surgery scheduling}
\label{sec:schedulinglit}

Surgery scheduling is arguably one of the most studied problems in healthcare operations research \citep{belien2024}.
We provide a brief overview of the relevant literature on the subject, focusing on studies that have dealt with surgical patient admission scheduling.
For a broader review of the literature on surgery scheduling, we refer the interested reader to the surveys by \citet{cardoen2010}, \citet{guerriero2010}, and \citet{zhu2019}.

Decisions concerning \acp{opr} are generally addressed at three decision levels: strategic, tactical, and operational -- both offline and online \citep{hulshof2012}.
Strategic decisions include determining the location, number, type, and opening hours of \acp{opr}, as well as setting appropriate staffing levels.
In tactical planning, several structural problems can be found, including the design of cyclic \ac{opr} timetables, commonly known as the \acf{mss} problem \citep{gupta2008}.
The \ac{mss} allocates available \ac{opr} capacity to surgeons or surgical disciplines according to their specific requirements.
Finally, operational surgery scheduling includes two types of problems: \textit{advance} scheduling and \textit{allocation} scheduling \citep{agnetis2012}.
The former is defined as the problem of setting a date and an \ac{opr} for each surgery.
In this regard, \citet{aringhieri2015} addressed an advance scheduling problem considering the allocation of \ac{opr} capacity to both surgical disciplines and (subsets of) patients.
In allocation scheduling problems, the assignment of a patient to a surgeon and a specific day is fixed, with the goal of sequencing procedures.
In addition, many studies have addressed the integration of advance and allocation scheduling \citep{riise2011,vanhuele2014,vancroonenburg2015,marques2019}.
In this context, when uncertainty is considered, it is often attributed to the duration of surgeries \citep{denton2007}.
For example, \citet{lamiri2008} used stochastic modeling to schedule both elective and emergency patients.
We refer to \citet{shehadeh2022} for a comprehensive survey of stochastic approaches for elective surgery scheduling with downstream capacity constraints.


Many studies have focused on integrating the operational planning stage with other resources, either concurrently \citep{bargetto2023} or in upstream and downstream stages \citep{harper2002}.
In tactical scheduling, constraints on both up-and downstream resources are often considered while generating the \ac{mss} \citep{fugener2014}.
For example, \citet{belien2009} used integer programming to create an \ac{mss} alongside the nurses' roster, while \citet{guido2017} considered capacity constraints of various downstream resources, including the post anesthesia center unit and \ac{icu} beds.
\citet{kianfar2023} included constraints on bed availability when generating a set of Pareto-efficient \acp{mss}.
At the operational decision level, \citet{bai2022} integrated surgery sequencing with recovery room planning.
Similarly, \citet{schneider2020} grouped patients based on resource utilization to account for different downstream resources during admission scheduling.
\cite{vancroonenburg2015} instead considered concurrent resources such as surgical staff and equipment, along with general dependencies between them and the surgeries.


\subsection{\ac{los} prediction}
\label{sec:lospredictionlit}

Several studies have used \ac{ml} to predict patients' \acp{los} \citep{faddy2009}.
As already mentioned, there is no consensus on which predictive approach is the best one, nor indications on which method to deploy in which context and how.
We refer the interested reader to \citet{stone2022} and \citet{bacchi2021} for two critical reviews of several papers using \ac{ml} techniques for \ac{los} prediction \new{and to \citet{garber2024} for a work comparing the operational efficiency of several \ac{ml} models}.

In the context of the present study, \citet{bacchi2022} identified two possible perspectives that can be adopted for \ac{los} prediction: \textit{classification} and \textit{regression}.
In classification, an \ac{ml} model is used to predict whether \ac{los} will be above or below a certain threshold -- e.g., $<7$ or $\geq7$ days \citep{davis1993} --, or to classify them in arbitrary duration bins -- e.g., 1-4 days, 5-8 days, or $\geq$12 days \citep{tsai2016}.
Regression models instead predict a scalar value, which is then used as the estimated \ac{los}, either left as-is or rounded to the nearest integer.
This perspective is the one more closely related to the framework are proposing, and will thus be the focus of the remainder of this section.

In \Cref{tab:regr}, we compare regression models reviewed by both \citet{bacchi2022} and \citet{stone2022}, focusing on studies that \textit{i}) propose data-driven methods for \ac{los} prediction and \textit{ii}) report model error metrics.
The studies are listed in ascending order of publication year and alphabetically by the first author's surname.
Based on available data, we report the \ac{los} of the studied populations using mean values ($\mu$), standard deviations ($\mu\pm\sigma$), intervals ($[min; max]$), or a combination of these.
Similarly, various error metrics are reported, including \acf{rmse}, \acf{mae}, \acf{mse}, and \acf{rmsle}.
In some cases, \ac{rmse} and \ac{mae} are reported in normalized form relative to their respective \acp{los}.
We refer to \citet{hyndman2006} for a comprehensive review of these metrics.

\begin{table}
\small
\centering
\begin{tabular}{llllll}
\hline
\textbf{Reference} & \textbf{Year} & \textbf{Population} & \textbf{\ac{los}} & \textbf{Metric} & \textbf{Performance} \\ 
\hline
\citet{tu1993} & \citeyear{tu1993} & Open heart surgery, \ac{icu} && \ac{rmse} & 0.056 \\
\citet{grigsby1994} & \citeyear{grigsby1994} & Orthopedics & $13.25\pm6.2$ & n\ac{mae} & 0.089 \\
&&&& n\ac{rmse} & 0.127 \\
\citet{mobley1995} & \citeyear{mobley1995} & Coronary care & $3.49 \; [1,20]$ & \ac{mae} & 1.43 \\
\citet{tremblay2006} & \citeyear{tremblay2006} & Digestive surgery & $[1;20]$ & \ac{mae} & 1.67-4.51 \\
\citet{liu2010} & \citeyear{liu2010} & Inpatients & $4.5 \pm 7.7$ & \acs{mse} & 29 \\ 
\citet{yang2010} & \citeyear{yang2010} & Burns & $22.85 \pm 20.7$ & \ac{mae} & 8.992-9.532 \\ 
\citet{huang2013} & \citeyear{huang2013} & Respiratory infections & $13.6 \; [2; 52]$ & \ac{rmse} & 1.75-8 \\
\citet{caetano2015} & \citeyear{caetano2015} & Inpatients && \ac{mae} & 0.224 \\
 &&&& \ac{rmse} & 0.469 \\
\citet{tsai2016} & \citeyear{tsai2016} & Heart failure & $8.24 \pm 5.87$ & \ac{mae} & 3.87-3.97 \\
&& Acute myocardial infarction & $6.97 \pm 5.95$ & \\
&& Coronary atherosclerosis & $2.63 \pm 2.25$ & \ac{mae} & 1.00-1.09 \\
\citet{turgeman2017} & \citeyear{turgeman2017} & Heart failure & $6.24 \pm 8.48$ & \ac{mae} & 1 \\
\citet{baek2018} & \citeyear{baek2018} & Inpatients & $7.01 \pm 9.82$ & \ac{mae} & 4.68 \\
\citet{cui2018} & \citeyear{cui2018} & Inpatients && \ac{rmse} & 3.10 \\
 &&&& \ac{mae} & 2.19 \\
\citet{daghistani2019} & \citeyear{daghistani2019} & Cardiology && \ac{mae} & 1.79 \\
 &&&& \ac{rmse} & 0.31 \\ 
\citet{muhlestein2019} & \citeyear{muhlestein2019} & Brain tumor surgery & $7.8\pm8.7$ & RMSLE & 0.631 \\
\citet{stone2020} & \citeyear{stone2020} & \acs{ed} & $[0; 200]$ & \ac{rmse} & 11.34 \\
\citet{boffmedeiros2023} & \citeyear{boffmedeiros2023} & Pediatric && \ac{mae} & 3.51 \\
\hline
\end{tabular}
\caption{Data on studies using regression to predict \ac{los}, as collected by \citet{bacchi2022} and \citet{stone2022}. Abbreviations: [n]\ac{rmse}: [normalized] root mean square error; [n]\ac{mae}: [normalized] mean absolute error; \acs{mse}: mean square error; \ac{rmsle}: root mean square logarithmic error.}
\label{tab:regr}
\end{table}

The earliest study by \citet{tu1993} trained a neural network to predict \ac{los} in the \ac{icu} following cardiac surgery.
Specific post-surgery \ac{los} was also the focus of \citet{tremblay2006} for digestive surgery patients and of \citet{muhlestein2019} for neurosurgery patients treated for brain tumors.
However, most studies listed in \Cref{tab:regr} focus on general inpatient \ac{los}.
\citet{yang2010} analyzed \ac{los} for burn patients, while \citet{stone2020} examined \ac{los} related to \acf{ed} visits, in both cases regardless of the intervention performed.
In contrast, several studies have focused on single surgical disciplines, grouping surgical and non-surgical cases, such as \citet{grigsby1994} in orthopedics and \citet{daghistani2019} in cardiology.
Similarly, \citet{mobley1995} and \citet{turgeman2017} analyzed and predicted \ac{los} in coronary care and heart failure patients, respectively.
\citet{tsai2016} analyzed three cardiological sub-populations, considering heart failure, acute myocardial infarction, and coronary atherosclerosis patients.
\citet{caetano2015}, \citet{baek2018}, and \citet{cui2018} dealt with diverse patient mixes, broadly considering data from all inpatients.
Moreover, \citet{liu2010} focused on a mix of surgical and non-surgical patients, coming from both waiting lists for elective procedures and from the \ac{ed}.
\citet{boffmedeiros2023} focused their analysis on pediatric patients, without considering the reason for their admission.
Finally, \citet{huang2013} explicitly considered non-surgical cases, analyzing the \ac{los} of patients admitted because of respiratory infections.
The wide range of disciplines interested in \ac{los} prediction reinforces the clinical significance of this topic.

Concerning the \acp{los} themselves, their distributions vary widely in terms of both average values -- from 3.49 days \citep{mobley1995} to 22.85 \citep{yang2010} -- and variability.
The lowest reported standard deviations among the various patient populations is 2.25 days \citep{tsai2016}, while the highest is 20.7 \citep{tsai2016}.
Such a wide range is not surprising, as different reasons for admission imply heterogeneous recovery processes.
Likewise, achieved predictive performance varies considerably, with \acp{rmse} ranging from 0.056 \citep{tu1993} to 11.34 \citep{stone2020} and \acp{mae} from 0.224 \citep{caetano2015} to 9.532 \citep{yang2010}.
Clearly, variations in error metrics can be partly attributed to differences in observed \acp{los} themselves, but they also underscore how researchers have been trying to predict \ac{los} values with considerably different outcomes.


\section{Problem description}
\label{sec:losproblem}

The problem setting under consideration involves two planning phases: a \emph{scheduling phase,} conducted weekly, and a \emph{rescheduling phase}, carried out daily except on Saturdays and Sundays.

The scheduling phase creates the admission schedule for the upcoming week by selecting a subset of patients from the waiting list and assigning each of them a surgery date and an \ac{opr}.
We assume that this phase takes place each Friday before the start of the following week.
Since the exact patients' \acp{los} are not yet known during the scheduling phase, a predicted value is used\footnote{We do not consider uncertainty with regards to surgery duration since we assume to be dealing with elective patients who are admitted for standardized procedures.}.
The true \ac{los} becomes known only after surgery, once the clinical evaluation of its success and the impact on the patient's health can be assessed.
Since the true \ac{los} may differ from the predicted value, changes to the admission schedule may be needed to prevent ward overcrowding.
Therefore, a rescheduling problem is solved on each weekday to repair any infeasibilities that may occur due to discrepancies between actual and predicted \ac{los} used in the scheduling phase.

The following sections present these two phases separately.


\subsection{Scheduling phase}
\label{ss:scheduling}

We consider a set of patients $P$ currently on a waiting list for elective surgery, who also require hospitalization in an inpatient ward for post-surgery recovery.
The required surgical procedures belong to a set $S$ of surgical disciplines.

Each patient $p \in P$ is characterized by their required surgical procedure, their latest possible day of surgery $f_p$\footnote{We assume that surgery always takes place on the first day of admission; hence, we use the terms ``day of surgery'' and ``day of admission'' interchangeably.}, the number of days they already spent in the waiting list $w_p$, and their urgency.
The urgency coefficient of patient $p$ is computed as $\pi_p = 360 / r_p$, where $r_p$ denotes the patient's maximum acceptable waiting time.
A higher $r_p$ results in a lower $\pi_p$, indicating lower urgency for that patient.
The surgical procedure required by patient $p$ is characterized by a surgery duration $u_p$ and an expected \ac{los} $l_p$.

We consider a set of wards $W$, each with a specific bed capacity $b_w$.
Due to specialized equipment and skill requirements, patients typically cannot be assigned to every ward.
However, we assume that each patient is compatible with at least one ward, and potentially multiple ones.
Let $P_w \subseteq P$ be the set of patients who can be admitted to ward $w$ and $W_p \subseteq W$ the set of wards to which patient $p$ can be admitted.

Surgeries are performed in a set of \acp{opr} $J$.
The \ac{mss} for all \acp{opr} is given and defines the amount $q_{sdj}$ of available time in \ac{opr} $j$ for each surgical discipline $s$ on each day $d$.

At the beginning of the planning period $D$, we assume that some patients who were admitted earlier are still recovering in the wards, occupying $h_{wd}$ beds in ward $w$ and day $d$.

The scheduling phase selects patients for surgery in the current planning period $D$ and assigns them to compatible wards while respecting bed capacity constraints.
Each selected patient must also be assigned an admission/surgery date while respecting \ac{opr} availability and capacity constraints.
The goal is to treat as many patients as possible during the current planning period, considering their urgency and the time they spent on the waiting list.
To achieve this, we minimize a simplified version of the objective function proposed by \citet{addis2016}, which employs a weighted sum of patient waiting times and surgery lateness.

We solve the scheduling problem as a \acf{milp} problem, with the relevant sets, parameters, and decision variables summarized in \Cref{tab:notationshort}.

\begin{table}
\centering
\small
\begin{tabular}{lp{0.8\textwidth}}
\hline
\multicolumn{2}{l}{\textbf{Sets}} \\
\hline
$P = \{ 1,...,\mathcal{P} \}$ & Set of patients on the waiting list, indexed by $p$ \\
$S = \{ 1,...,\mathcal{S} \}$ & Set of surgical disciplines, indexed by $s$ \\
$W = \{ 1,...,\mathcal{W} \}$ & Set of wards in the hospital, indexed by $w$ \\
$D = \{ 1,...,\mathcal{D} \}$ & Set of days in the current planning period, indexed by $d$ \\
$J = \{ 1,...,\mathcal{J} \}$ & Set of \acp{opr}, indexed by $j$ \\
$P_w \subseteq P$ & Set of patients that can be admitted to ward $w$ based on their surgical discipline \\
$W_p \subseteq W$ & Set of wards to which patient $p$ can be admitted to based on their surgical discipline \\
\hline
\multicolumn{2}{l}{\textbf{Parameters}} \\
\hline
$f_p \in D$ & Surgery due date of patient $p$ \\
$w_p \geq 0$ & Time spent on the waiting list by patient $p$ at the start of the current planning period \\
$\pi_p \geq 0$ & Urgency coefficient of patient $p$ \\
$l_p > 0$ & Expected \ac{los} of patient $p$, in days \\	
$u_p \geq 0$ & Surgery duration of patient $p$, in minutes \\
$b_{w} > 0$ & Number of beds available in ward $w$ \\
$q_{sdj} \geq 0$ & Available time in \ac{opr} $j$ for surgical discipline $s$ on day $d$, in minutes \\
$h_{wd} \geq 0$ & Number of beds occupied in ward $w$ on day $d$ by previously admitted patients \\
\hline
\multicolumn{2}{l}{\textbf{Decision variables}} \\
\hline
$x_{pwdj} \in \{0, 1\}$ & Binary variable, equal to 1 if patient $p$ is assigned to ward $w$ on day $d$ and \ac{opr} $j$ and 0 otherwise \\
$\delta_p \geq 0$ & Non-negative continuous variable, equal to the lateness of patient $p$ with respect to their due date $f_p$ \\
$\omega_p \geq 0$ & Non-negative continuous variable, equal to the time spent on the waiting list by patient $p$ in the current planning period \\
\hline
\end{tabular}
\caption{Sets, parameters, and decision variables of the scheduling model.}
\label{tab:notationshort}
\end{table}

For each $p \in P$, $w \in W_p$, $d \in D$, and $j \in J$, let $x_{pwdj}$ be a binary variable that takes value 1 if patient $p$ is admitted to ward $w$ on day $d$ while being operated on in \ac{opr} $j$, and 0 otherwise.
For each $p \in P$, the continuous non-negative variable $\omega_p$ counts how many days patient $p$ has spent on the waiting list, including the days before the beginning of the scheduling period.
Similarly, the continuous non-negative variable $\delta_p$ keeps track of lateness with respect to the surgery due date $f_p$ of patient $p$.

The scheduling problem is formulated as follows:
\begin{equation}
\min \, \sum_{p\in P} \pi_p \left( \delta_p + \omega_p \right)
\label{sec:obj}
\end{equation}
s.t.
\begin{align}
& \sum_{w \in W_{p}}\sum_{d \in D}\sum_{j \in J} x_{pwdj} \leq 1 & \forall p \in P
\label{sec:atmostonce} \\
& \sum_{p \in P}\sum_{w \in W_p} u_p\cdot x_{pwdj} \leq q_{sdj} & \forall s \in S, d \in D, j \in J
\label{sec:or_capacity} \\
& \sum_{j \in J}\sum_{p \in P_w}\sum^{d}_{\substack{d^\prime= \max{\{0;d-l_p+1\}}}} x_{pwd^{\prime}j} \leq b_{w}-h_{dw} & \forall w \in W, d \in D
\label{sec:ward_capacity} \\
& \delta_p \geq \sum_{d \in D}\sum_{w \in W_p}\sum_{j \in J} \left( d\cdot x_{pwdj} \right) - f_p & \forall p \in P
\label{sec:lateness} \\
& \delta_p \geq |D|\cdot\left(1-\sum_{d \in D}\sum_{w \in W_p}\sum_{j \in J}x_{pwdj}\right)-f_p & \forall p \in P
\label{sec:lateness2} \\
& \omega_p \geq \sum_{d \in D}\sum_{w \in W_p}\sum_{j \in J} (d\cdot x_{pwdj}) + w_p & \forall p \in P
\label{sec:waiting} \\
& \omega_p \geq |D|\cdot\left(1-\sum_{d \in D}\sum_{w \in W_p}\sum_{j \in J}x_{pwdj}\right) + w_p & \forall p \in P
\label{sec:waiting2} \\
& x_{pwdj} \in \{ 0,1 \} & \forall p \in P, w \in W_p, d \in D, j \in J
\label{sec:x_existance} \\
& \delta_p \geq 0 & \forall p \in P
\label{sec:delta_existance} \\
&\omega_p \geq 0 & \forall p \in P
\label{sec:omega_existance}
\end{align}

Objective function \eqref{sec:obj} minimizes the weighted sum of patients' total waiting time and surgery lateness, with weights reflecting patient urgency.
Constraints \eqref{sec:atmostonce} ensure that each patient is assigned at most one surgery date, \ac{opr}, and compatible ward.
Constraints \eqref{sec:or_capacity} ensure the \ac{opr} availability and capacity for each surgical discipline is never exceeded.
Constraints \eqref{sec:ward_capacity} ensure that the bed capacity of the wards is not exceeded.
Constraints \eqref{sec:lateness} and \eqref{sec:lateness2} calculate the lateness of patient $p$ with respect to their surgery due date $f_p$, depending on whether $p$ undergoes surgery or not.
Constraints \eqref{sec:waiting} and \eqref{sec:waiting2} determine the waiting time of patient $p$ in the current planning period.
Finally, Constraints \eqref{sec:x_existance}-\eqref{sec:omega_existance} enforce bounds on the decision variables.
Although $\delta_p$ and $\omega_p$ are defined as continuous variables, the constraints in which they appear, along with the objective function, effectively restrict their values to integers.

\subsection{Rescheduling phase}

Because of errors in the predicted \ac{los}, solutions generated in the scheduling phase may become infeasible when the patients' real \acp{los} are revealed.
To repair these infeasibilities, a rescheduling problem is solved on each weekday which adjusts the admission schedule for the remaining days in the current planning period.
More specifically, the rescheduling phase aims to find a feasible solution given the updated \ac{los} while minimizing deviations from admission dates and ward assignments determined during the scheduling phase.
We consider the following four rescheduling policies:
\begin{enumerate}
\item \textbf{Postpone admissions (P)} -- Admit patients later than the admission date determined during the scheduling phase.
\item \textbf{Change the admission ward (CW)} -- Admit patients to a different compatible ward than the one assigned during the scheduling phase.
\item \textbf{Transfer patients between wards (T)} -- Change the ward of patients that are already hospitalized, transferring them to a different compatible ward.
\item \textbf{Combined policy (C)} -- A combination of the first three policies that allows to postpone admissions, change admission wards, and transfer hospitalized patients.
\end{enumerate}

When deploying a rescheduling policy on day $d^*$, the following additional sets and parameters are introduced.
Let $\hat{D} = \{ d^*,...,D \}$ be the remaining days in the planning period.
Let $\hat{P}(d^*) \subseteq P$ denote the set of patients who, during the scheduling phase, were assigned an admission date on or after $d^*$.
Among them, let $\hat{P}_w(d^*)$ represent those eligible for admission to ward $w$.
For each patient $p \in \hat{P}(d^*)$, let $\hat{d}_p$ be the admission date assigned in the scheduling phase, $\hat{w}_p$ the assigned ward, and $\hat{j}_p$ the assigned \ac{opr}.


Additionally, $\hat{P}(d^*)^H \subseteq P$ denotes the subset of patients who have already been admitted to the hospital before day $d^*$.
Only a subset $\hat{P}(d^*)^{HT}$ of these patients are eligible for transfers, i.e., patients whose condition is not critical and who have not already been transferred during their stay.
During rescheduling, these patients must always be part of the schedule -- they can never be removed from the schedule.
We do so by setting their admission date to the first day of the rescheduling planning period $\hat{d}_p = d^*$, setting their surgery duration $u_p$ to zero and updating their estimated \ac{los} based on how much time they have already spent in the hospital, thereby considering only the resources they are yet to consume.

We implement the rescheduling policies by solving the corresponding rescheduling problems as \ac{milp} problems.
The primary decision variable in all the rescheduling \ac{milp} models is the same as the assignment variable in the scheduling model.
For each $p \in \hat{P}(d^*)$, $w \in W$, $d \in \hat{D}$, and $j \in J$, let $x_{pwdj}$ be a binary variable that takes value 1 if, after the rescheduling, patient $p$ is admitted to ward $w$ and \ac{opr} $j$ on day $d$, and 0 otherwise.

We use four additional variables to keep track of whether or not, and how, patient $p$ is affected by the rescheduling.
Let $y_p$ be a binary variable that takes value 1 if the admission date of patient $p \in \hat{P}(d^*)$ is changed, and 0 otherwise.
Let $v_p$ be a binary variable that takes value 1 if the ward to which $p \in \hat{P}(d^*)$ is admitted changes in the rescheduling phase, and 0 otherwise.
Let $\chi_p$ be a binary variable that takes value 1 if patient $p \in \hat{P}(d^*)^{HT}$ is transferred to a different ward during their stay, and 0 otherwise.
Finally, let $z_p$ be a binary variable that takes value 1 if patient $p \in \hat{P}(d^*)$ is canceled, that is, if they are no longer part of the updated admission schedule, and 0 otherwise.

\Cref{tab:notationshortrescheduling} provides an overview of the additional sets, parameters, and decision variables used in the \ac{milp} models of the rescheduling policies.

\begin{table}
\centering
\small
\begin{tabular}{lp{0.75\textwidth}}
\hline
\multicolumn{2}{l}{\textbf{Sets}} \\
\hline
$\hat{D}=\{d^*,..., |D|\}$ & Set of days in the rescheduling planning period \\
$\hat{P}(d^*) \subseteq P $ & Subset of patients whose admission dates are greater than or equal to $d^*$ \\ 
$\hat{P}(d^*)^H \subseteq P$ & Subset of patients admitted before day $d^*$ who are still in the hospital \\
$\hat{P}(d^*)^{HT} \subseteq \hat{P}(d^*)^H$ & Subset of patients admitted to the hospital before day $d^*$ that can be transferred to another ward \\
$\hat{P}_w(d^*) \subseteq \hat{P}(d^*)$ & Subset of patients whose admission dates are greater than or equal to $d^*$ that can be admitted to ward $w$ \\
\hline
\multicolumn{2}{l}{\textbf{Parameters}} \\
\hline
$d^* \in D$ & Day in the original planning period on which rescheduling occurs \\
$\hat{d}_p \in D$ & Admission day assigned to patient $p$ in the scheduling phase \\
$\hat{j}_p \in J$ & \ac{opr} assigned to patient $p$ in the scheduling phase \\
$\hat{w}_p \in W$ & Ward assigned to patient $p$ in the scheduling phase \\
\hline
\multicolumn{2}{l}{\textbf{Decision variables}} \\
\hline
$x_{pwdj} \in \{0,1\}$ & Binary variable equal to 1 if patient $p$ is assigned to ward $w$ on day $d$ and \ac{opr} $j$, 0 otherwise \\
$z_p \in \{0,1\}$ & Binary variable equal to 1 if patient $p$ is excluded from the schedule in the rescheduling phase, 0 otherwise \\
$y_p \in \{0,1\}$ & Binary variable equal to 1 if the admission date of patient $p$ is postponed, 0 otherwise \\
$v_p \in \{0,1\}$ & Binary variable equal to 1 if the ward to which patient $p$ is admitted is changed, 0 otherwise \\
$\chi_p \in \{0,1\}$ & Binary variable equal to 1 if patient $p$ is transferred to a different ward, 0 otherwise \\ 
\hline
\end{tabular}
\caption{Additional sets, parameters, and decision variables of the rescheduling policies.}
\label{tab:notationshortrescheduling}
\end{table}

\subsubsection{Postpone patient admissions (P)}

The rescheduling problem in which patient admissions can be postponed, while the ward and OT are kept as decided by the scheduling phase, is formulated as follows:
\begin{equation}
\min \, \sum_{p \in \hat{P}(d^*)} y_p + \left( |\hat{P}(d^*)|-|\hat{P}(d^*)^H| \right) \sum_{p \in \hat{P}(d^*)} z_p
\label{r1:obj}
\end{equation}
s.t.
\begin{align}
& x_{p\hat{w}_pd^{*}\hat{j}_p} = 1 & \forall p \in \hat{P}(d^*)^H
\label{r1:c1a} \\
& x_{p\hat{w}_pd^{*}\hat{j}_p} = 1 & \forall p \in \hat{P}(d^*) : \hat{d}_p = d^{*}
\label{r1:c1b} \\ 
& \sum_{d \in \hat{D}: d \geq \hat{d}_p}\sum_{j \in J} x_{p\hat{w}_pdj} \leq 1 & \forall p \in \hat{P}(d^*) : \hat{d}_p > d^{*}
\label{r1:atmostonce_a} \\ 
& \displaystyle\sum\limits_{p \in \hat{P}(d^*)} u_p x_{p\hat{w}_pdj} \leq q_{sdj} & \forall s \in S, d \in \hat{D}, j \in J
\label{r1:c3} \\
& \sum_{j \in J}\sum_{\substack{p \in \hat{P}_w(d^*): \hat{w}_p = w}}\sum^{d}_{\substack{d^\prime= \max{\{0;d-l_p+1\}}}} x_{p\hat{w}_pd^{\prime}j} \leq b_{w}-h_{dw} & \forall w \in W, d \in \hat{D}
\label{r1:c4} \\
& y_p \geq 1 - \sum_{j\in J}x_{p\hat{w}_p\hat{d}_{p}j} & \forall p \in \hat{P}(d^*)
\label{r1:c5} \\
& z_p \geq 1-\displaystyle\sum\limits_{d \in \hat{D}}\displaystyle\sum\limits_{j \in J} x_{p\hat{w}_pdj} & \forall p \in \hat{P}(d^*) 
\label{r1:c2} \\
& x_{p\hat{w}_pdj} \in \{ 0,1 \} & \forall p \in \hat{P}(d^*), d \in \hat{D}, j \in J
\label{r1:c6} \\
& y_p \in \{ 0,1 \} & \forall p \in \hat{P}(d^*)
\label{r1:c7} \\
& z_p \in \{ 0,1 \} & \forall p \in \hat{P}(d^*)
\label{r1:c8}
\end{align}

Objective function \eqref{r1:obj} minimizes the number of patients whose admission date is changed and the number of canceled patients.
The latter is weighted by $|\hat{P}(d^*)|-|\hat{P}(d^*)^H|$ to ensure that patients are preferably postponed rather than canceled.

Constraints \eqref{r1:c1a} and \eqref{r1:c1b} ensure that patients already present in the hospital before or on day $d^*$ are included in the solution.
Constraints \eqref{r1:atmostonce_a}-\eqref{r1:c4} are equivalent to Constraints \eqref{sec:atmostonce}-\eqref{sec:ward_capacity} in the scheduling model, with the appropriate indices.
Constraints \eqref{r1:c5} link $y_p$ and $x_{pwdj}$ variables, while Constraints \eqref{r1:c2} link $z_p$ and $x_{pwdj}$ variables.
Finally, Constraints \eqref{r1:c6}-\eqref{r1:c8} enforce bounds on the decision variables.

\subsubsection{Change the admission ward (CW)}

The rescheduling problem describing the possibility of a patient admitted after $d^*$ being admitted to a different ward than the one originally scheduled, while keeping the same \ac{opr} and day $\hat{d}_p$, is formulated as follows:
\begin{equation}
\min \, \sum_{p \in \hat{P}(d^*)} v_p + \left( |\hat{P}(d^*)|-|\hat{P}(d^*)^H| \right) \sum_{p \in \hat{P}(d^*)} z_p
\label{r2:obj}
\end{equation}
s.t.
\begin{align}
& x_{p\hat{w}_pd^*\hat{j}_p} = 1 & \forall p \in \hat{P}(d^*)^H 
\label{r2:c1a} \\
& x_{p\hat{w}_pd^*\hat{j}_p} = 1 & \forall p \in \hat{P}(d^*) : \hat{d}_p = d^{*} 
\label{r2:c1b} \\ 
& \sum_{w \in W_p}\sum_{j \in J} x_{pw\hat{d}_pj} \leq 1 & \forall p \in \hat{P}(d^*) : \hat{d}_p > d^{*}
\label{r2:atmostonce} \\ 
& \displaystyle\sum\limits_{p \in \hat{P}(d^*)}\sum\limits_{w \in W_p} u_p x_{pw\hat{d}_pj} \leq q_{sdj} & \forall s \in S, d \in \hat{D}, j \in J
\label{r2:c3} \\
& \sum_{j \in J}\sum^{d}_{\substack{d^\prime= \max{\{0;d-l_p+1\}}}}\sum_{\substack{p \in \hat{P}_w(d^*): \hat{d}_p = d^{\prime}}} x_{pwd^{\prime}j} \leq b_{w}-h_{dw} & \forall w \in W, d \in \hat{D}
\label{r2:c4} \\
& v_p \geq 1 - \sum_{j\in J}x_{p\hat{w}_p\hat{d}_pj} & \forall p \in \hat{P}(d^*)
\label{r2:c5} \\
& z_p \geq 1 - \displaystyle\sum\limits_{w \in W_p}\displaystyle\sum\limits_{j \in J} x_{pw\hat{d}_pj} & \forall p \in \hat{P}(d^*)
\label{r2:c2} \\ 
& x_{pw\hat{d}_pj} \in \{ 0,1 \} & \forall p \in \hat{P}(d^*), w \in W_p, j \in J
\label{r2:c6} \\
& v_p \in \{ 0,1 \} & \forall p \in \hat{P}(d^*)
\label{r2:c7} \\
& z_p \in \{ 0,1 \} & \forall p \in \hat{P}(d^*) 
\label{r2:c8}
\end{align}

Objective function \eqref{r2:obj} minimizes the number of patients whose ward is changed and the number of patients that cannot be rescheduled.
As before, the latter term is weighted by $|\hat{P}(d^*)|-|\hat{P}(d^*)^H|$ to prioritize ward changes over patient cancellations.

Constraints \eqref{r2:c1a} and \eqref{r2:c1b} ensure that patients admitted before or on day $d^*$ are included in the solution.
Constraints \eqref{r2:atmostonce}-\eqref{r2:c4} are equivalent to Constraints \eqref{sec:atmostonce}-\eqref{sec:ward_capacity}.
Constraints \eqref{r2:c5} link $v_p$ and $x_{pwdj}$ variables, while Constraints \eqref{r2:c2} link $z_p$ and $x_{pwdj}$ variables.
Finally, Constraints \eqref{r2:c6}-\eqref{r2:c8} enforce bounds on the decision variables.

\subsubsection{Transfer patients between wards (T)}

The rescheduling problem defined by the policy in which already admitted patients can be transferred between wards at most once during their stay is formulated as follows:
\begin{equation}
\min \, \sum_{p \in \hat{P}(d^*)} v_p + \sum\limits_{p \in \hat{P}(d^*)^{HT}} \chi_p + \left( |\hat{P}(d^*)|-|\hat{P}(d^*)^H| \right) \sum_{p \in \hat{P}(d^*)} z_p
\label{r3:obj}
\end{equation}
s.t.
\begin{align}
& x_{p \hat{w}_pd^*\hat{j}_p}=1 & \forall p \in \hat{P}(d^*)\setminus \hat{P}(d^*)^{HT}
\label{r3:c0} \\
& \sum_{w \in W_p}x_{pwd^*\hat{j}_p} = 1 & \forall p \in \hat{P}(d^*)^{HT}
\label{r3:c1a} \\
& \sum_{w \in W_p} x_{pwd^{*}\hat{j}_p} = 1 & \forall p \in \hat{P}(d^*) : \hat{d}_p = d^{*}
\label{r3:c1b} \\ 
& \sum_{w \in W_p}\sum_{j \in J} x_{pw\hat{d}_pj} \leq 1 & \forall p \in \hat{P}(d^*) : \hat{d}_p > d^{*} 
\label{r3:atmostonce} \\ 
& \displaystyle\sum\limits_{p \in \hat{P}_{sd}(d^*)}\sum\limits_{w \in W_p} u_p x_{pw\hat{d}_pj} \leq q_{sdj} & \forall s \in S, d \in \hat{D}, j \in J 
\label{r3:c3} \\
& \sum_{j \in J}\sum^{d}_{\substack{d^\prime= \\\max{\{0;d-l_p+1\}}}}\sum_{\substack{p \in \hat{P}_w(d^*): \\ \hat{d}_p = d^{\prime}}} x_{pwd^{\prime}j} \leq b_{w}-h_{dw} & \forall w \in W, d \in \hat{D} 
\label{r3:c4} \\
& v_p \geq 1 - \sum_{j\in J}x_{p\hat{w}_{p}\hat{d}_pj} & \forall p \in \hat{P}(d^*) 
\label{r3:c5a} \\
& \chi_p \geq 1 - x_{p\hat{w}_{p}\hat{d}_p\hat{j}_p} & \forall p \in \hat{P}(d^*)^{HT}
\label{r3:c5b} \\
& z_p \geq 1 - \displaystyle\sum\limits_{w \in W_p}\displaystyle\sum\limits_{j \in J} x_{pw\hat{d}_pj} & \forall p \in \hat{P}(d^*) 
\label{r3:c2} \\ 
& x_{pw\hat{d}_pj} \in \{ 0,1 \} & \forall p \in \hat{P}(d^*), w \in W_p, j \in J 
\label{r3:c6} \\
& v_p \in \{ 0,1 \} & \forall p \in \hat{P}(d^*) 
\label{r3:c7} \\
& \chi_p \in \{ 0,1 \} & \forall p \in \hat{P}(d^*)^{HT}
\label{r3:c8} \\
& z_p \in \{ 0,1 \} & \forall p \in \hat{P}(d^*) 
\label{r3:c9}
\end{align}

Objective function \eqref{r3:obj} minimizes the number of patients whose ward is changed during their stay and the number of canceled patients, weighted by $|\hat{P}(d^*)|-|\hat{P}(d^*)^H|$.

Constraints \eqref{r3:c0} ensure that patients in the hospital before day $d^*$ who cannot be transferred are indeed not transferred.
Constraints \eqref{r3:c1a} and \eqref{r3:c1b} ensure that admitted patients are included in the solution, without fixing the ward they were originally assigned to.
Constraints \eqref{r3:atmostonce}-\eqref{r3:c4} are equivalent to Constraints \eqref{sec:atmostonce}-\eqref{sec:ward_capacity}.
Constraints \eqref{r3:c5a}, \eqref{r3:c5b} and \eqref{r3:c2} link $x_{pwdj}$ variables with $v_p$, $\chi_p$, and $z_p$, respectively.
Finally, Constraints \eqref{r3:c6}-\eqref{r3:c9} enforce bounds on the decision variables.

\subsubsection{Combined policy (C)}


The rescheduling problem defining the combined policy is formulated as follows:
\begin{equation}
\min \, \sum_{p \in \hat{P}(d^*)} (y_p + v_p) + \sum\limits_{p \in \hat{P}(d^*)^{HT}} \chi_p + \left( |\hat{P}(d^*)|-|\hat{P}(d^*)^H| \right) \sum_{p \in \hat{P}(d^*)} z_p
\label{r4:obj}
\end{equation}
s.t.
\begin{align}
& x_{p \hat{w}_pd^*\hat{j}_p}=1 & \forall p \in \hat{P}(d^*)\setminus \hat{P}(d^*)^{HT}
\label{r4:c0} \\
& \sum_{w \in W_p}x_{pwd^*\hat{j}_p} = 1 & \forall p \in \hat{P}(d^*)^{HT}
\label{r4:c1a} \\
& \sum_{w \in W_p} x_{pwd^{*}\hat{j}_p} = 1 & \forall p \in \hat{P}(d^*) : \hat{d}_p = d^{*}
\label{r4:c1b} \\ 
& \sum_{d \in \hat{D}:d \geq \hat{d}_p}\sum_{w \in W_p}\sum_{j \in J} x_{pwdj} \leq 1 & \forall p \in \hat{P}(d^*) : d > d^{*}
\label{r4:atmostonce} \\ 
& \sum\limits_{p \in \hat{P}_{sd}(d^*)}\sum\limits_{w \in W_p} u_p x_{pw\hat{d}_pj} \leq q_{sdj} & \forall s \in S, d \in \hat{D}, j \in J 
\label{r4:c3} \\
& \sum_{j \in J}\sum^{d}_{\substack{d^\prime= \max{\{0;d-l_p+1\}}}}\sum_{p \in \hat{P}_w(d^*)} x_{pwd^{\prime}j} \leq b_{w}-h_{dw} & \forall w \in W, d \in \hat{D}
\label{r4:c4} \\
& y_p \geq 1 - \sum\limits_{w \in W_p}\sum_{j\in J}x_{p\hat{w}_p\hat{d}_{p}j} & \forall p \in \hat{P}(d^*) 
\label{r4:c5} \\
& v_p \geq 1 - \sum_{d \in \hat{D}}\sum_{j\in J}x_{p\hat{w}_{p}\hat{d}_pj} & \forall p \in \hat{P}(d^*) 
\label{r4:c6a} \\
& \chi_p \geq 1 - x_{p\hat{w}_{p}\hat{d}_p\hat{j}_p} & \forall p \in \hat{P}(d^*)^{HT} 
\label{r4:c6b} \\
& z_p \geq 1 - \sum_{d \in \hat{D}}\sum\limits_{w \in W_p}\displaystyle\sum\limits_{j \in J} x_{pw\hat{d}_pj} & \forall p \in \hat{P}(d^*)
\label{r4:c7} \\ 
& x_{pwdj} \in \{ 0,1 \} & \forall p \in \hat{P}(d^*), w \in W_p, d \in \hat{D}, j \in J
\label{r4:c8} \\
& y_p \in \{ 0,1 \} & \forall p \in \hat{P}(d^*) 
\label{r4:c9} \\
& v_p \in \{ 0,1 \} & \forall p \in \hat{P}(d^*) 
\label{r4:c11} \\
& \chi_p \in \{ 0,1 \} & \forall p \in \hat{P}(d^*)^{HT} 
\label{r4:c10} \\
& z_p \in \{ 0,1 \} & \forall p \in \hat{P}(d^*) 
\label{r4:c12}
\end{align}

Objective function \eqref{r4:obj} minimizes the number of patients whose admission is postponed, who are admitted to a different ward than the one they were originally assigned to, or whose ward is changed during their stay, and the number of patients that cannot be rescheduled.
Cancellations are once again penalized by $|\hat{P}(d^*)|-|\hat{P}(d^*)^H|$, while postponements, admission changes, and transfers all have the same unitary relative weight.

Constraints \eqref{r4:c0} ensure that patients in the hospital before day $d^*$ who cannot be transferred are indeed not transferred.
Constraints \eqref{r4:c1a} and \eqref{r4:c1b} ensure that the admitted patients are included in the solution, without fixing the ward they were originally assigned to.
Constraints \eqref{r4:atmostonce}, \eqref{r4:c3}, and \eqref{r4:c4} are equivalent to Constraints \eqref{sec:atmostonce}-\eqref{sec:ward_capacity}.
Constraints \eqref{r4:c5}, \eqref{r4:c6a}, \eqref{r4:c6b}, and \eqref{r4:c7} link $x_{pwd}$ variables with $y_p$, $v_p$, $\chi_p$, and $z_p$, respectively.
Finally, Constraints \eqref{r4:c8}-\eqref{r4:c12} enforce bounds on the decision variables.

\section{Simulated predictions}
\label{sec:losprediction}

As described in \Cref{sec:lospredictionlit}, \ac{ml} model performance for regression is evaluated based on its predictive error, regardless of the specific metric used.
If a model is always correct, with zero error, its error distribution corresponds to a Dirac delta function.
In practice, however, \ac{ml} models are not perfect oracles and are instead characterized by a non-zero error metric.
Generally, models are trained and tested by minimizing prediction error, aiming to reduce the difference between true and predicted values as much as possible.
Common error metrics, such as \ac{rmse} or \ac{mae}, treat errors symmetrically, penalizing under- and overestimations equally during model training and validation.
This is done to obtain an \textit{unbiased} predictive model whose predictions are, on average, correct.
Consequently, the error distribution of an unbiased predictor can be approximated as symmetric around zero, reflecting the absence of systematic over- or underestimation.
Hence, any symmetric distribution with a mean of zero can be used to approximate the error from an unbiased predictor, such as a normal distribution with $\mu=0$.
Interestingly, the Dirac delta function that can be used to model the error distribution of a perfect oracle can be interpreted as the limit of a normal distribution with mean $\mu=0$ when its standard deviation $\sigma$ approaches zero:
\begin{equation}
\delta(x)=\lim_{\sigma\rightarrow 0}\frac{e^{-\frac{1}{2}\left(\frac{x}{\sigma}\right)^2}}{\sigma\sqrt{2\pi}}
\end{equation}

When $\sigma\neq0 $, this function can be used to represent varying error degrees, proportional to the variability in the performance of a predictive model.
Adopting this perspective allows us to represent unbiased \ac{ml} predictive models through their errors, modeled using symmetric probability distributions.
For simplicity, we approximate prediction errors with normal distributions, although alternatives such as symmetric triangular or Cauchy-Lorentz distributions are also viable.

However, not all \ac{ml} models are unbiased. 
In some cases, systematic under- or overestimations occur, requiring error distributions that reflect this bias. 
This can be modeled by shifting the Dirac delta function as follows:
\begin{equation}
 \delta(x) = \begin{cases} 0, & x + c \neq 0 \\ {\infty} , & x + c= 0 \end{cases}
 \label{shifted_delta}
\end{equation}
where $c$ is a quantity representative of the bias.
Any Dirac delta function defined by Equation \eqref{shifted_delta} represents the error produced by a model that consistently deviates from the true value by exactly $c$.
We can therefore model biased errors using a normal distribution with $\mu \neq 0$.

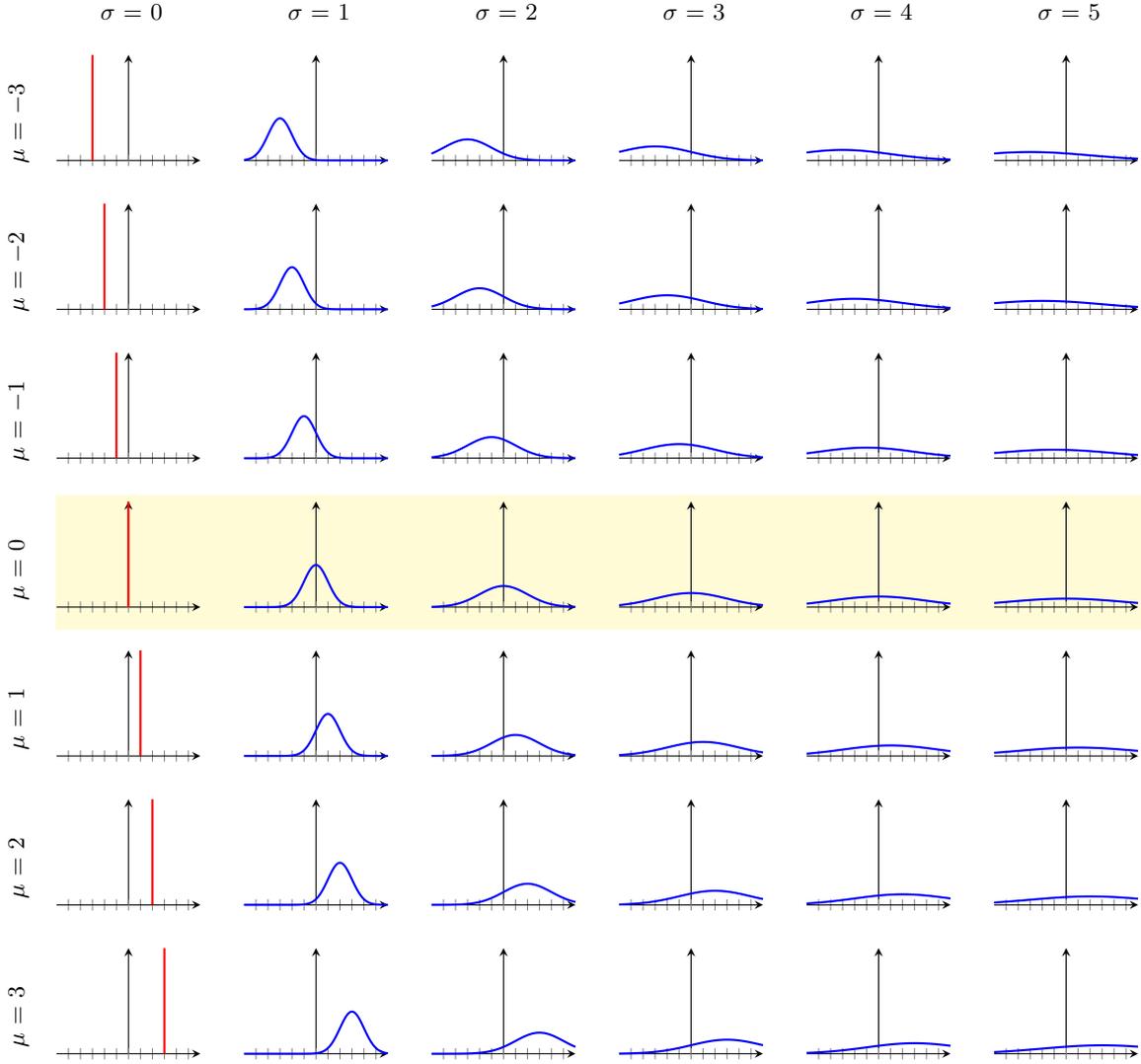
\begin{figure}[h!]
\centering
\begin{tikzpicture}
\pgfmathsetmacro\highlightrow{-3.25} 
\fill[yellow!20] (0, \highlightrow*2-0.8) rectangle (14.5, \highlightrow*2+1);
\foreach \col in {0,...,5} {
 \pgfmathsetmacro\stddev{\col}
 \node at (\col*2.5 + 1, 1) {\footnotesize $\sigma$ = \stddev};
}
\foreach \row in {0,...,6} {
 \pgfmathsetmacro\mean{-3 + \row}
 \node[rotate=90] at (-.5, -\row*2 - 0.5) {\footnotesize $\mu$ = \pgfmathprintnumber[precision=0]{\mean}};
}
\foreach \row in {0,...,6} {
 \foreach \col in {0,...,5} {
 \pgfmathsetmacro\mean{-3 + \row}
 \pgfmathsetmacro\stddev{\col}
 \begin{scope}[shift={(\col*2.5, -(1+\row*2))}]
 \begin{axis}[
 width=3.5cm, height=3cm,
 axis x line=middle, axis y line=middle,
 ymin=0, ymax=1,
 xmin=-6, xmax=6,
 xtick={-5, -4, -3, -2, -1, 0, 1, 2, 3, 4, 5}, 
 xticklabels={},
 extra x ticks={0}, 
 extra x tick labels={}, 
 ytick={0},
 samples=100,
 domain=-6:6,
 clip=false
 ]
 \ifdim \stddev pt = 0pt
 \addplot[red, thick, ycomb] coordinates {(\mean,1)};
 \else
 \addplot[blue, thick] {1/(sqrt(2*pi)*\stddev) * exp(-0.5*((x-\mean)/\stddev)^2)};
 \fi
 \end{axis}
 \end{scope}
 }
}
\end{tikzpicture}
\caption{Plots of error distributions characterized by mean $\mu$ and standard deviation $\sigma$ as normal distributions (or Dirac delta functions, when $\sigma = 0$). The error distributions of unbiased predictors are highlighted in yellow.}
\label{fig:errordistr}
\end{figure}
\Cref{fig:errordistr} shows the prediction error distributions we consider in this paper.
We argue that these distributions reflect the performance typically observed in models discussed in the literature.
The distributions corresponding to unbiased predictors, which represent the most realistic and desirable cases, are highlighted in yellow.
To simulate \ac{los} predictions, we add errors drawn from the error distributions to the patients' true \acp{los}.
Specifically, the predicted \ac{los} of patient $p$ is simulated as $l_p = \hat{l}_p + \epsilon$ with $\epsilon \sim \mathcal{N}(\mu, \sigma^2)$ and $\hat{l}_p$ the patient's true \ac{los}.

\section{Computational experiments}
\label{sec:losexpml}

To study the interaction between \ac{los} prediction accuracy and rescheduling, we conducted a computational study.
A dataset consisting of five problem instances was generated based on historical records of the largest university hospital in Belgium, following the procedure described by \citet{vancroonenburg2019}.
A scheduling horizon of eight weeks and three surgical disciplines were considered.
The maximum waiting time for each patient was determined by randomly selecting a value from the set \{8, 30, 60, 180, 360\} using the probabilities \{0.08, 0.37, 0.37, 0.17, 0.01\} \citep{valente2009model}.
For each instance, 42 \ac{los} prediction scenarios were considered by varying the error mean from -3 to 3 and the error standard deviation from 0 to 5
These values are representative of the error values reported in \Cref{tab:regr}, and thus reflect the predictive performance currently achieved in the literature.
These values are also sufficiently low that a prediction model with such performance could be considered for implementation in practice.
Larger errors would result in a model that is highly unreliable and deemed unsuitable for practical applications.

To account for the stochastic nature of the experiments, each experiment was repeated five times, resulting in a total of 1050 runs (5 instances $\times$ 42 \ac{los} prediction scenarios $\times$ 5 repetitions).
All experiments were carried out on an AMD Ryzen 9 5950X 16-core processor at 3.40 GHz with 64 GB of RAM.
The integer programming problems were solved using Gurobi 11.0.0 with the optimality gap set to $\text{e-}2$ and configured to use a single thread.
All code was written in Python, using the gurobipy library to interface with Gurobi.
The average runtime to solve each problem to the configured optimality gap was 32.3 seconds, with a maximum of 150.9 seconds and a minimum of 1.2 seconds.

\section{Results}
\label{sec:losresults}

This section presents the results of the computational experiments.
To evaluate the performance of each rescheduling policy as a function of varying \ac{los} prediction errors, we consider the following three types of metrics: objective function values (\Cref{res:obj}), patient-related metrics (\Cref{res:pat-related}), and ward occupancy rates (\Cref{res:occ-rate}).

\subsection{Objective values}
\label{res:obj}

\begin{figure}[h]
 \centering
 \includegraphics[width=.73\textwidth]{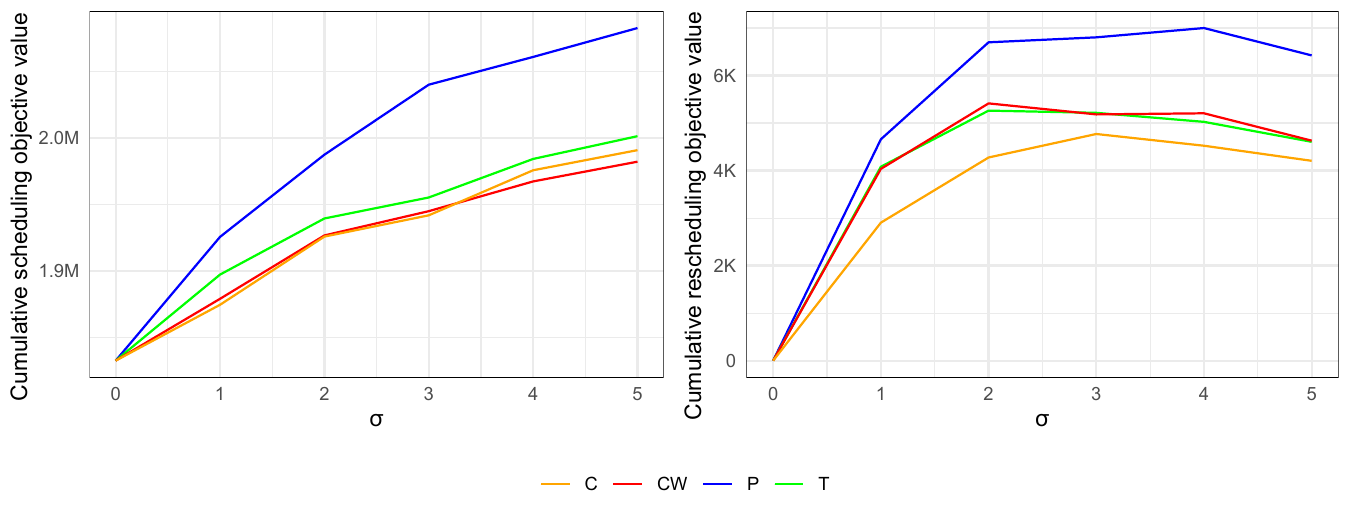}
 \caption{Sum of the objective function values of both the scheduling (left) and rescheduling (right) phases with unbiased predictions (i.e., $\mu=0$).}
 \label{fig:obj_mean0}
\end{figure}

The first metric is related to the objective functions of the scheduling and rescheduling problems.
\Cref{fig:obj_mean0} reports the cumulative objective values reached by the scheduling (left) and rescheduling (right) phases for an unbiased predictor.
The x-axis of both graphs shows the standard deviation $\sigma$ of the prediction error, while the y-axis shows the sum of the objective function values accumulated over the respective phases.
Recall that the objective function in the scheduling phase is defined as a weighted sum of the total number of days patients spend waiting, including any tardy days, while the objective function in the rescheduling phase sums the number of changes and the (penalized) number of cancellations.
The four rescheduling objective functions \eqref{r1:obj}, \eqref{r2:obj}, \eqref{r3:obj}, and \eqref{r4:obj} are defined such that they are mutually comparable.

The first insight that can be obtained is intuitive: in case of an unbiased prediction, having a lower $\sigma$ (and therefore a lower absolute error) provides the best results for both scheduling and rescheduling objective function values.
However, the plot on the right shows that increasing values of $\sigma$ do not lead to a monotonous increase in the rescheduling costs.
This suggests that, even if the predictive error cannot be reduced, its impact can still be mitigated by using rescheduling policies.
Moreover, assuming a symmetrical distribution of errors -- as is often the case in practice and as discussed above -- underestimates can sometimes balance overestimates and vice versa, further justifying this asymptotic behavior.

When comparing the policies, C outperforms the others in terms of the rescheduling objective values, while it performs comparably to CW and slightly better than T in the scheduling phase.
It can also be observed that P consistently yields the worst performance in both phases, which is reasonable as both waiting times and delays increase with postponement.

\begin{figure}[h!]
 \centering
 \begin{subfigure}{.8\textwidth}
 \centering
 \includegraphics[width=1\linewidth]{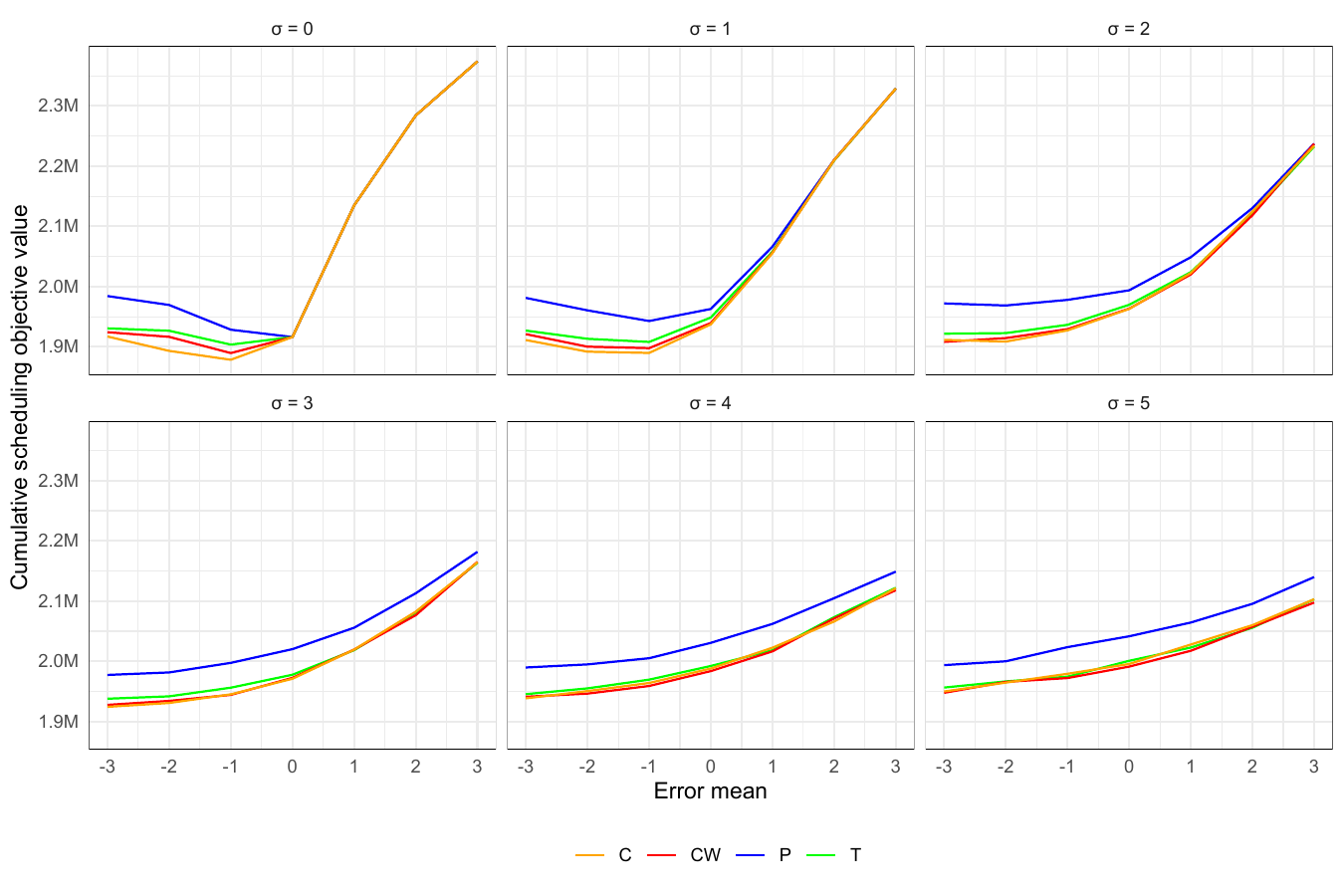}
 \caption{Sum of the objective values of the scheduling phases.}
 \label{fig:scheduling}
 \end{subfigure}
 \begin{subfigure}{.8\textwidth}
 \centering
 \includegraphics[width=1\linewidth]{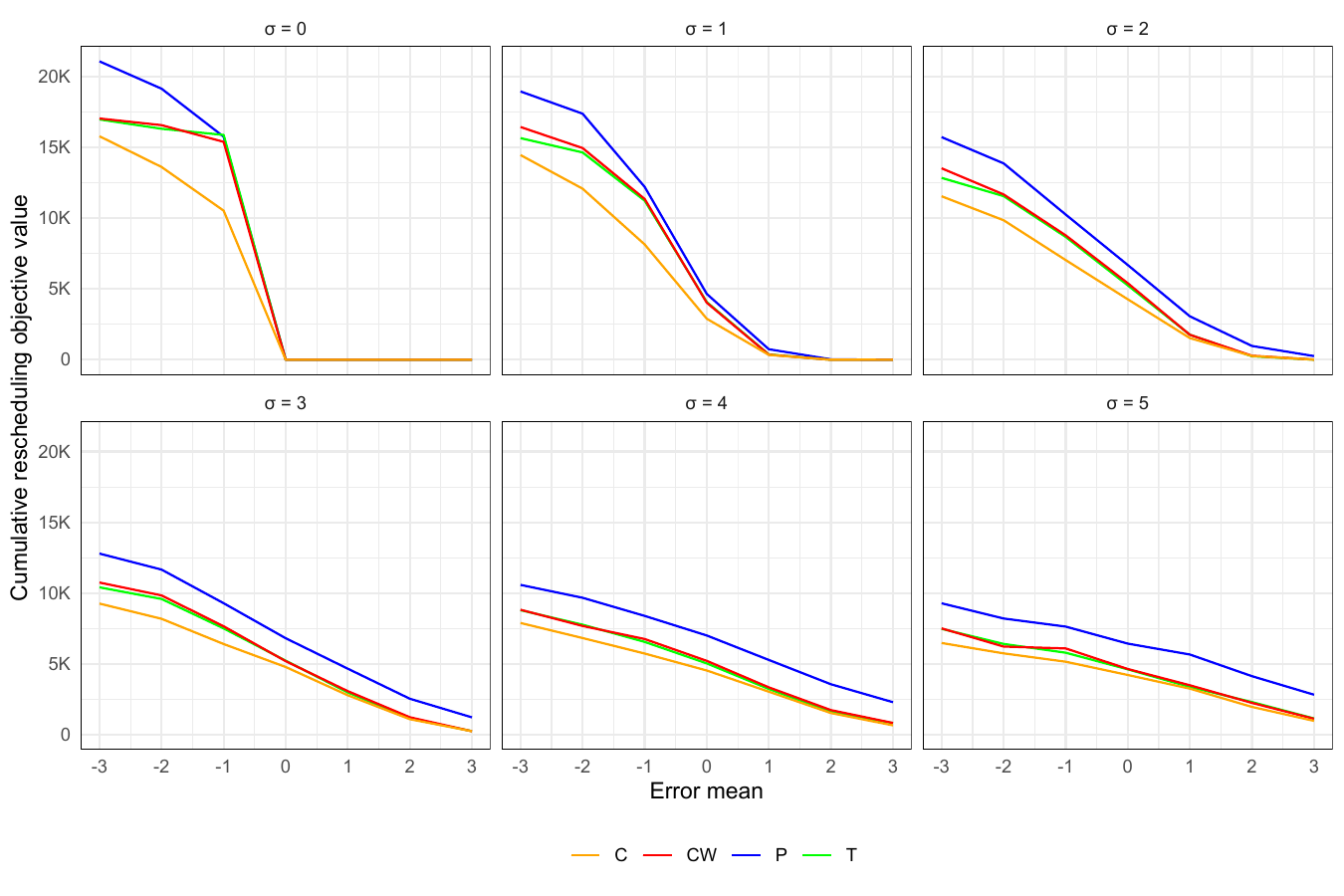}
 \caption{Sum of the objective values of the rescheduling phases.}
 \label{fig:rescheduling}
 \end{subfigure}
 \caption{Effect of error mean and standard deviation on the sum of objective values in the (a) scheduling and (b) rescheduling phases.}
 \label{fig:obj_values}
\end{figure}

\Cref{fig:obj_values} shows the cumulative scheduling and rescheduling objective values for the four rescheduling policies for both unbiased and biased predictors with varying error degree.
These two figures are organized as follows: each panel, starting from the top left, reports the results obtained with a fixed standard deviation of the simulated error, ranging from 0 to 5.
Within each panel, the x-axis shows the mean of the error, ranging from $-3$ to $3$.
In other words, $\mu<0$ represents scenarios in which the prediction error leads to an underestimation of \acp{los}, while $\mu > 0$ represents scenarios in which \acp{los} are overestimated.
$\mu=0$ coincides with plots in \Cref{fig:obj_mean0}.
The top-left panel shows the effects of consistent under- or overestimations, without any variability in the results.\\
As shown in \Cref{fig:scheduling}, underestimating \acp{los} generally leads to lower values of the cumulative scheduling objective function.
This occurs because patients are consistently assumed to require fewer resources than they actually need, leading to a higher number of scheduled patients.
On the other hand, overestimating patients' \acp{los} results in resource under-utilization, which creates backlogs in the waiting list and increases the total number of waiting days.
This trend becomes less pronounced as $\sigma$ increases, as larger $\sigma$ values reduce the impact of both under- and overestimations.
The increased standard deviation widens the range of potential errors between predicted and actual \acp{los}, making it more likely for the errors to change sign relative to the error mean, as illustrated in \Cref{fig:errordistr}.


\Cref{fig:rescheduling} demonstrates that the cumulative sum of the rescheduling objective values decreases as the mean of the prediction error increases.
Overestimating patients' \ac{los} produces less dense schedules, where little to no adjustment is needed to resolve infeasibilities, given that patients are assumed to require more resources than they actually do.
This effect is most pronounced when predictions are consistently biased ($\sigma=0$), resulting in a rescheduling objective value of 0 for all instances with $\mu \geq 0$.
As $\sigma$ increases, the impact of varying $\mu$ diminishes due to the error compensation phenomenon previously described.

From a policy perspective, policy P consistently underperforms compared to the other policies in terms of both scheduling and rescheduling objective values.
This is expected, as policy P addresses infeasibilities by postponing patient admissions, which naturally results in longer waiting times and, consequently, higher scheduling objective values.
Additionally, it fails to utilize any spare capacity that may become available during rescheduling cycles, leading to increased rescheduling objective values.
Policies T and CW perform similarly across almost all scenarios.
In cases of underestimation, the C policy clearly outperforms the others.

\subsection{Patient-related metrics}
\label{res:pat-related}

This section discusses detailed patient-related solution quality metrics.
These are partially related to the objective values discussed in \Cref{res:obj} but provide a more concrete representation of the outcomes resulting from the (re-)optimized schedules in practice.
While rescheduling policies can improve the system performance, they can cause inconveniences for both patients and hospital staff.
The metrics we consider in this section capture these inconveniences, offering additional insights into both policy performance and the effect of predictive errors.
The considered metrics are as follows:
\begin{enumerate}
\item The number of patients remaining on the waiting list at the end of the scheduling horizon;
\item The number of patients whose surgery has been canceled during a rescheduling phase;
\item The number of patients affected, but not canceled, by the rescheduling policies.
This includes patients who were postponed, had their admission ward changed, or were transferred between wards during their stay.
\end{enumerate}

Postponements are essentially free for hospitals, but they can be inconvenient for patients, as they may require adjustments to personal and family schedules.
In some cases, postponements may also have negative clinical consequences.
Transfers are also not comfortable for the patients and can be cumbersome for hospital staff as well.
Transferring a recovering patient requires both the sending and receiving wards to prepare by allocating personnel and resources.
Finally, changing the admission ward is the least disruptive policy for patients but may require additional effort from hospital staff, such as reorganizing shifts and tasks.
For example, nurses trained to care for patients recovering from a specific type of surgery may not work as efficiently with patients recovering from another type of surgery.

\begin{figure}[h]
 \centering
 \includegraphics[width=0.8\textwidth]{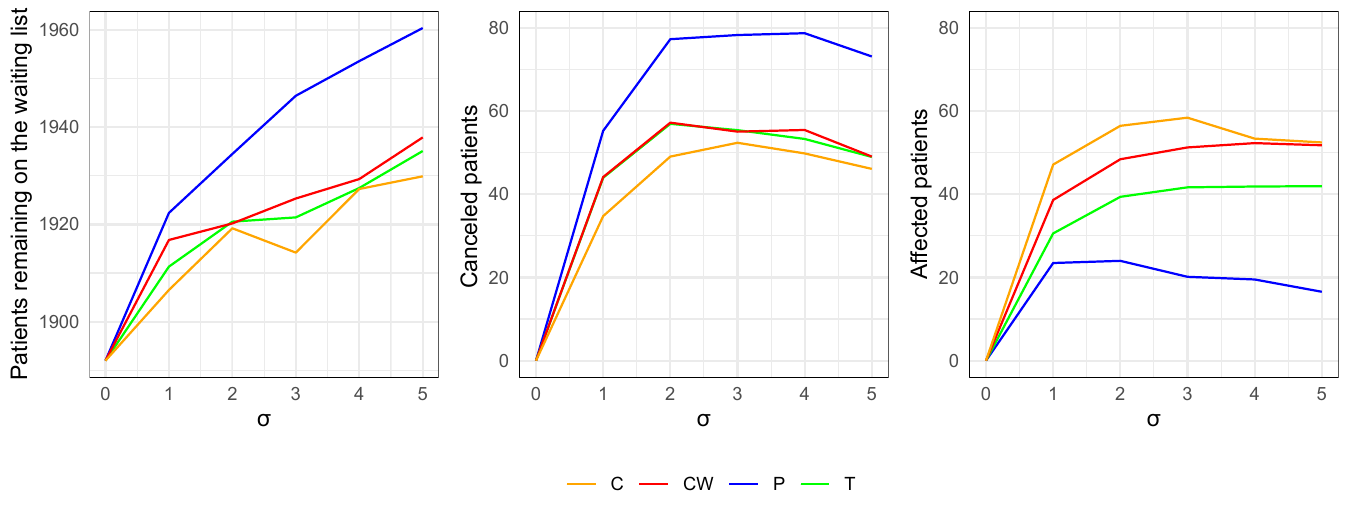}
 \caption{Number of patients on the waiting list at the end of the period (left), canceled patients (middle), and affected patients (right) for unbiased predictions.}
 \label{fig:metrics0}
\end{figure}

\Cref{fig:metrics0} reports all metrics for scenarios with unbiased predictions ($\mu=0$) and varying $\sigma$.
First, the difference between the smallest and the largest number of patients remaining on the waiting list is approximately 70.
From these patients' perspective, this implies having to be called for surgery further in the future, thus waiting longer, increasing their discomfort, and, for some, posing the risk of worsened health outcomes.
Unsurprisingly, the number of canceled patients follows the same trend of the cumulative sum of the objective values of the rescheduling phases (\Cref{fig:obj_mean0}, right).
This suggests that the primary driver of rescheduling objective function values is the cost of surgical cancellations, rather than the cost of the policies themselves.
At the same time, if we examine the number of affected patients, it is clear that policies have different performances.
Policy C, due to its ability to exploit the most degrees of flexibility, affects more patients compared to the other policies.
Policy P affects significantly fewer patients, indicating that schedules generated with an unbiased predictor offer limited opportunities for postponements.
Both CW and T policies show an intermediate performance, with a quasi-asymptotical behavior as $\sigma$ increases and with T affecting approximately 10 patients less than CW.

\begin{figure}[h!]
	\centering
 \begin{subfigure}{.9\textwidth}
		\centering
		\includegraphics[width=1\linewidth]{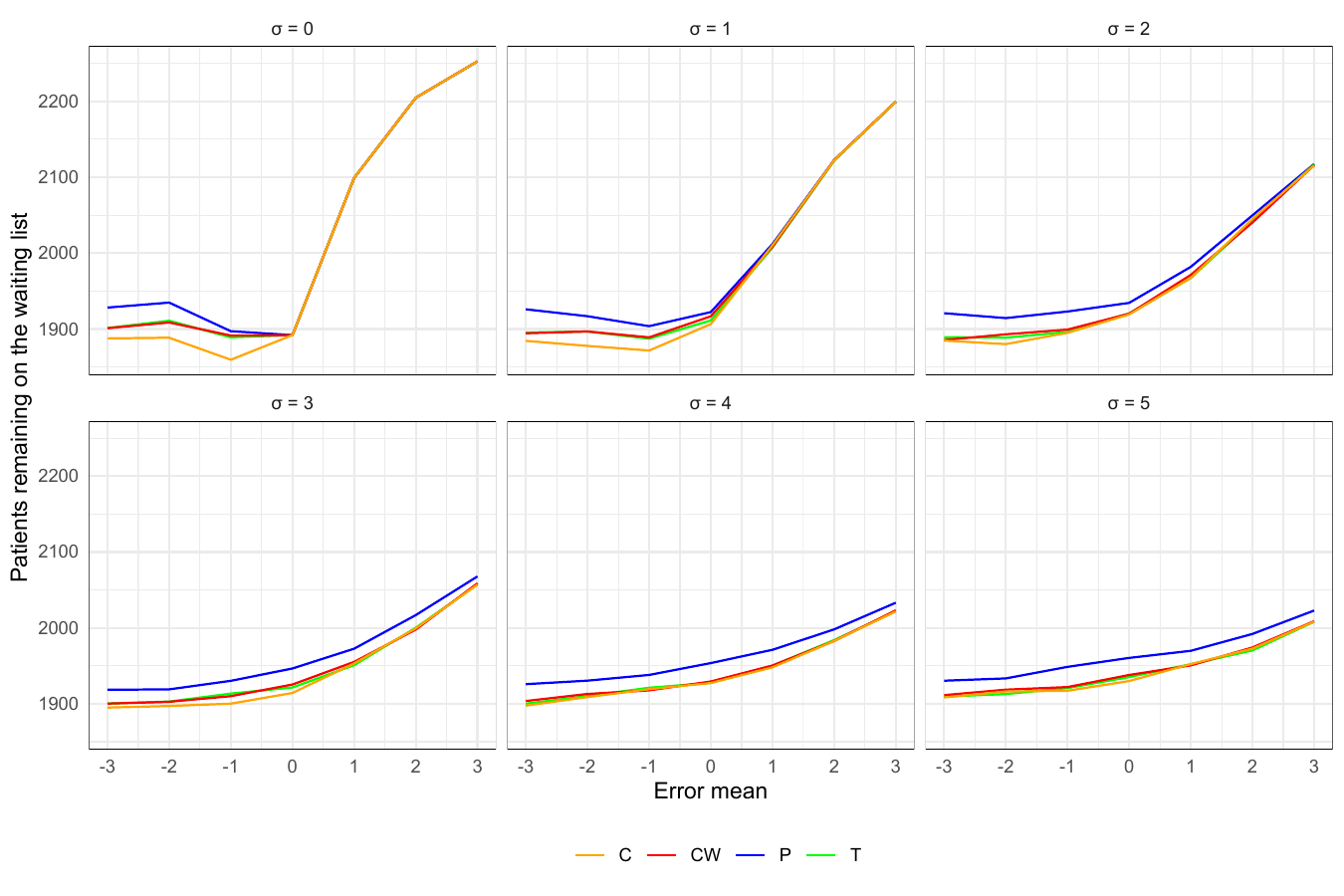}
		\caption{Number of patients on the waiting list.}
 \label{fig:wl}
	\end{subfigure}

 \begin{subfigure}{.9\textwidth}
		\centering
		\includegraphics[width=1\linewidth]{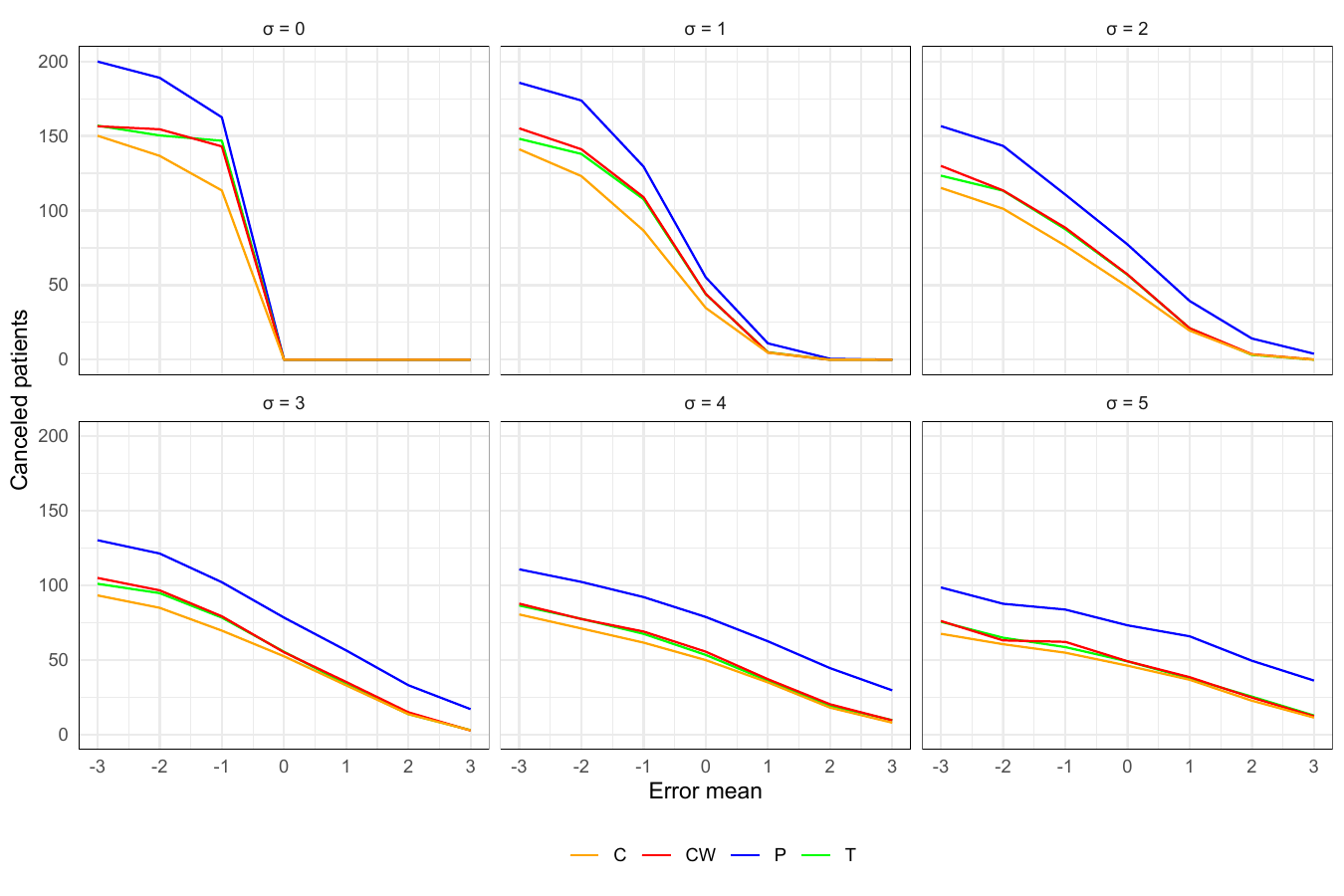}
		\caption{Number of canceled patients.}
 \label{fig:canc}
	\end{subfigure}
 \end{figure}
\begin{figure}\ContinuedFloat
\centering 
 \begin{subfigure}{.9\textwidth}
		\centering
		\includegraphics[width=1\linewidth]{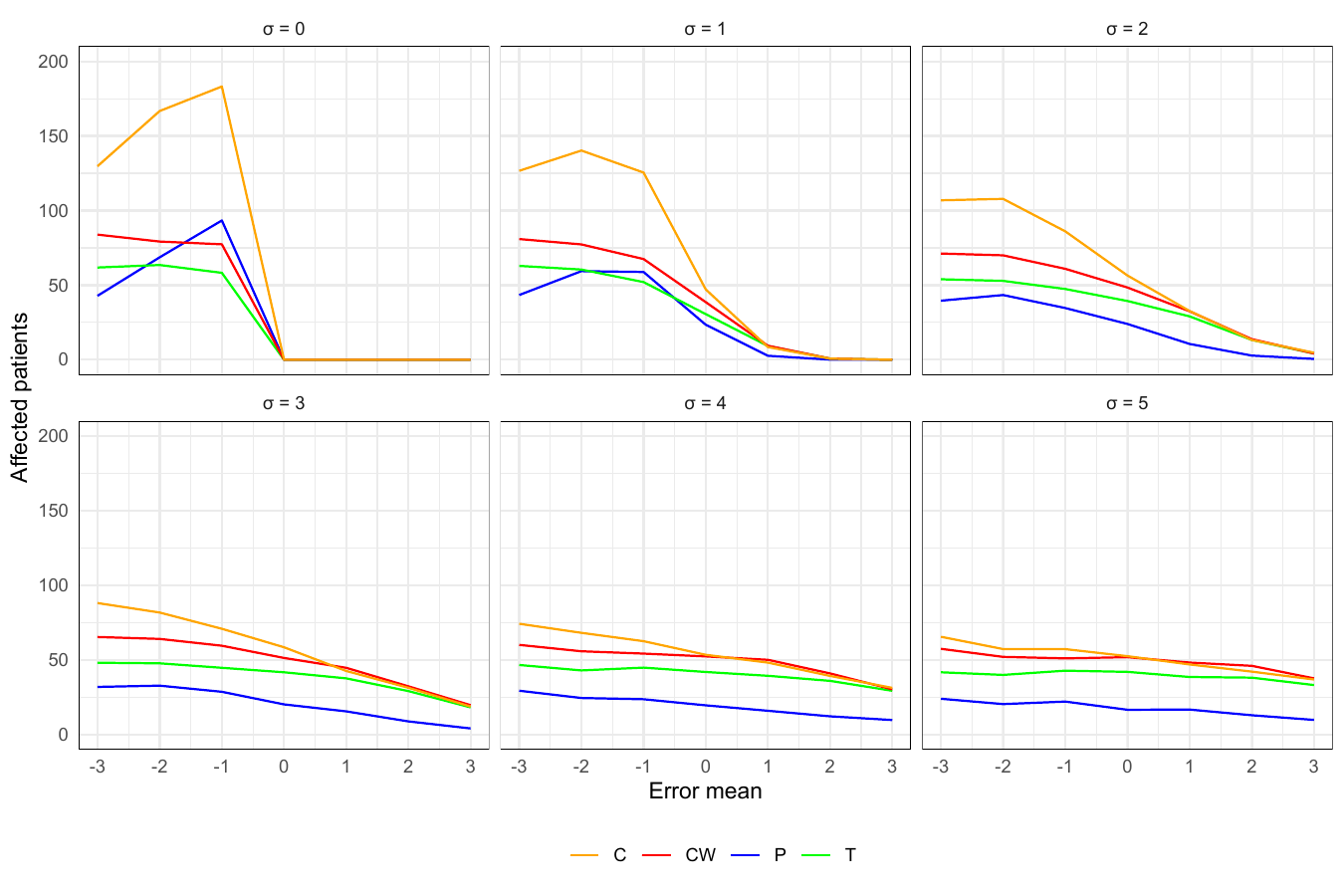}
		\caption{Number of affected patients.}
 \label{fig:aff}
	\end{subfigure}
	\caption{Effect of error mean and standard deviation on (a) number of patients on the waiting list, (b) canceled patients, and (c) number of affected patients.}
 \label{fig:metrics}
\end{figure}

\Cref{fig:metrics} displays the three patient-related metrics for the different rescheduling policies, considering both unbiased and biased predictors.
\Cref{fig:wl} plots the number of patients remaining on the waiting list at the end of the scheduling period, while \Cref{fig:canc} plots the total number of canceled patients.
Again, these two plots show the same trends as \Cref{fig:scheduling} and \Cref{fig:rescheduling}, respectively.
Examining Figures \ref{fig:wl} and \ref{fig:canc} together also allows to better assess the trade-off between under- and overestimation.
Underestimating patient \ac{los} results in fewer patients on the waiting list but also leads to an increased number of cancellations.
Overestimations, in contrast, increase the size of the waiting list but significantly reduce cancellations.
Regarding policy performance, the increased flexibility provided by policy C results in lower values for both metrics, while P performs the worst, particularly in terms of patient cancellations.
This suggests that simply postponing patients is not sufficient to prevent cancellations, namely because some patients are simply pushed out of the scheduling horizon.

The information presented in \Cref{fig:aff} provides additional insights that are not immediately evident from the analysis of objective function values alone.
These results demonstrate that overestimating \acp{los} with $\sigma=0$ requires no schedule adjustments, resulting in zero affected patients across all policies.
Moreover, as the standard deviation increases, the number of affected patients decreases, regardless of the considered error mean.

Finally, the minimum number of affected patients when $\sigma \neq 0$ does not occur at $\mu=0$ but rather at $\mu=3$.
This is again due to overestimation, which allocates more resources to patients than actually necessary.

In terms of policy performance, policy P affects the fewest patients, while C causes the most disruptions, particularly in scenarios with underestimation.
Finally, T and CW achieve an intermediate performance, both of them aligning closely with C when $\mu \geq 0$.
However, T affects less patients than CW, especially when $\mu\leq 0$.

\subsection{Ward occupancy rate}
\label{res:occ-rate}

\begin{figure}[h]
\centering
\includegraphics[width=0.8\textwidth]{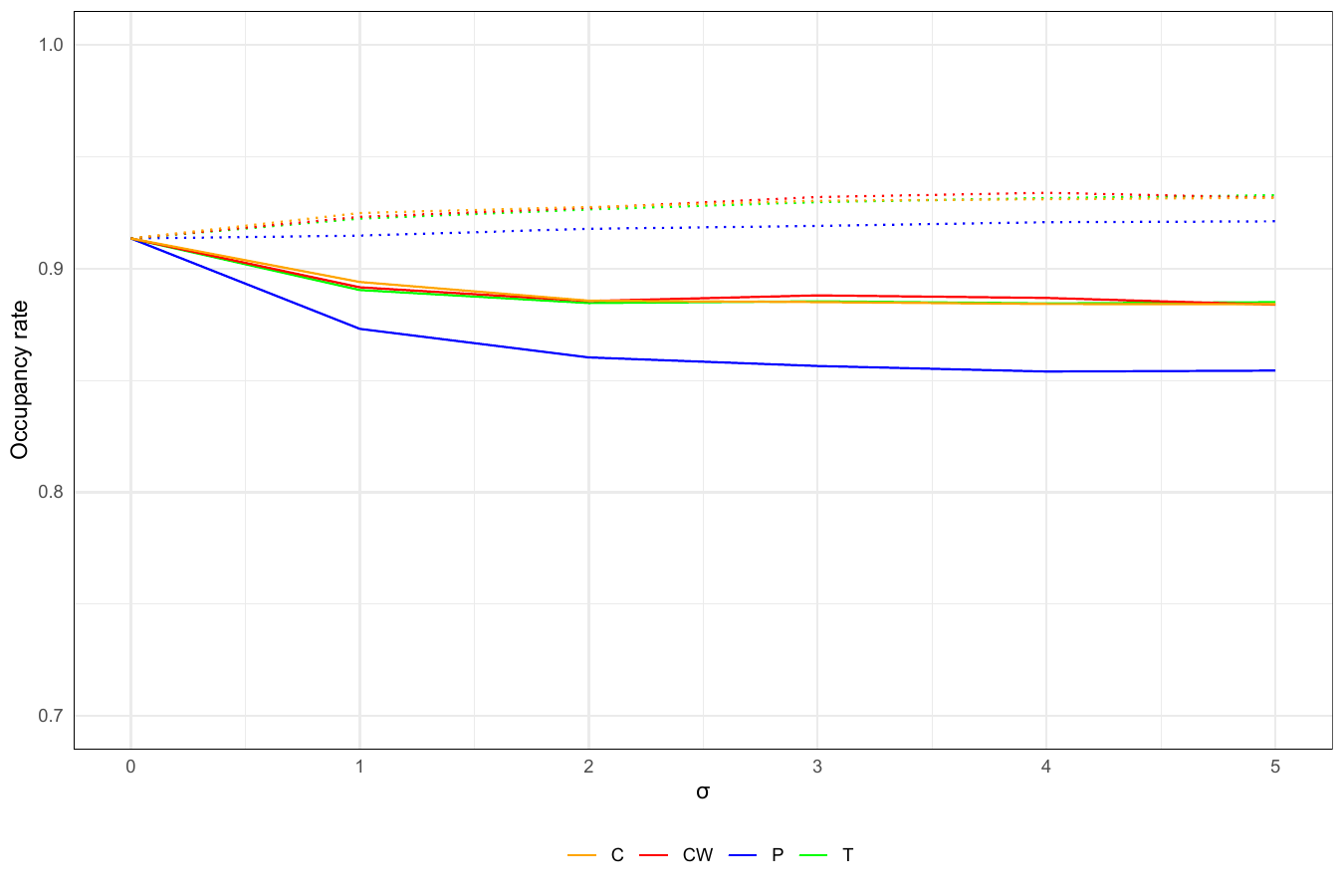}
\caption{Effect on ward occupancy rate for unbiased predictions. Dashed lines are the predicted occupancy rates, and solid lines are the observed ones.}
\label{fig:occupancy0}
\end{figure}

We now focus on analyzing how ward occupancy is affected by the prediction error and the different rescheduling policies.
\Cref{fig:occupancy0} shows the predicted and observed bed occupancy rates across all wards, using dashed lines for the occupancy in schedules computed using the predicted \acp{los} and solid lines for the occupancy computed using the observed (true) \acp{los}, all based on an unbiased predictor.
As expected, the highest observed occupancy rates occur when $\sigma=0$.
For any higher value of $\sigma$, the predicted and observed occupancy rates diverge by approximately 5\% for policies T, CW, and C and by approximately 7\% for P.
Furthermore, policy P consistently results in the worse values for this metric.
Interestingly, the difference between predicted and observed occupancy rates also shows a quasi-asymptotic behavior.
This implies that even if the prediction is not precise, there is a minimum theoretical occupancy rate that can be consistently achieved through the scheduling and rescheduling cycles.
Moreover, efforts to reduce the $\sigma$ of a \ac{los} predictor will have little impact on this metric unless $\sigma$ is reduced below 3.


\begin{figure}[h]
\centering
\includegraphics[width=0.8\textwidth]{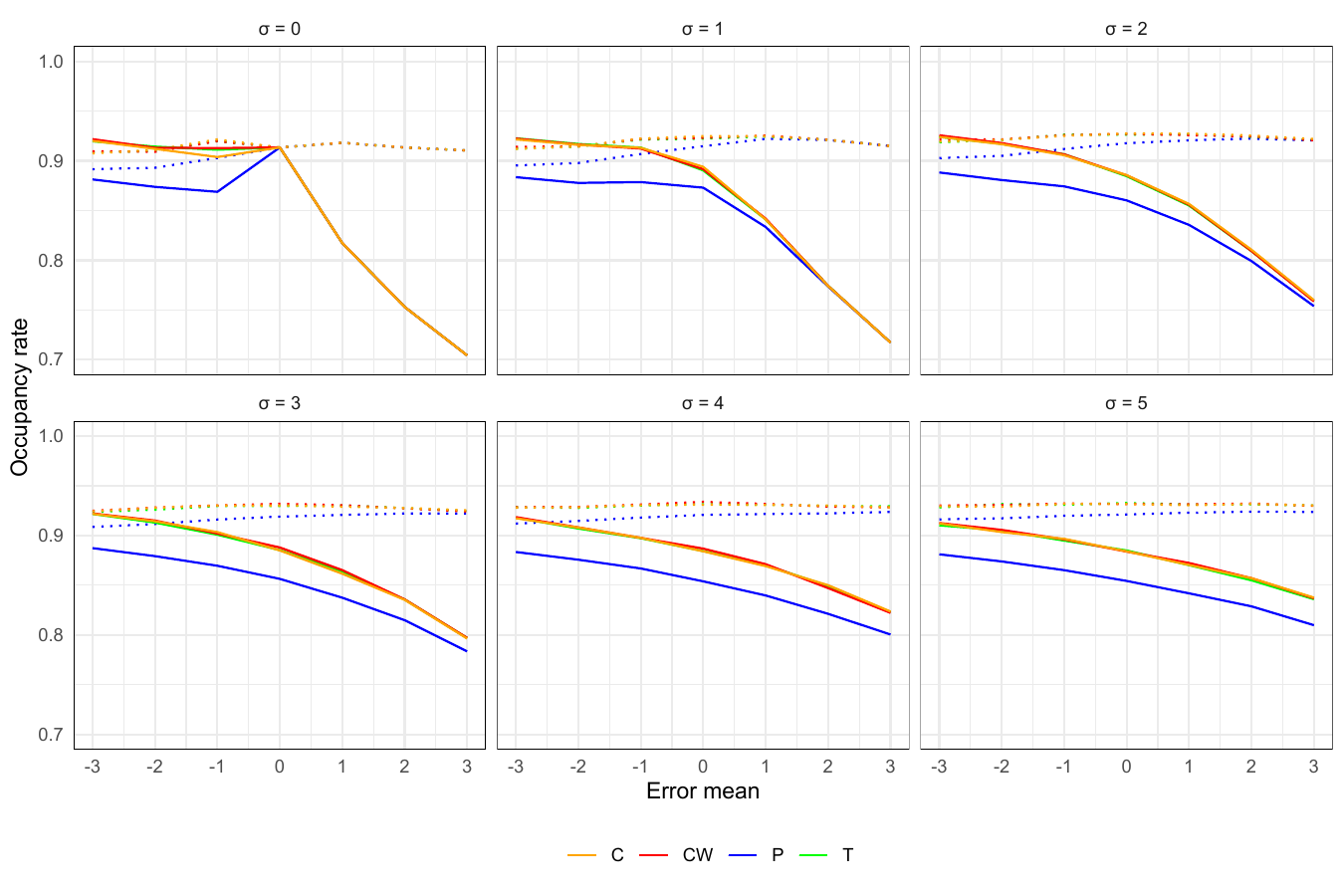}
\caption{Effect on occupancy rate. Solid lines are the predicted occupancy rates and dashed lines are the observed ones.}
\label{fig:occupancy}
\end{figure}

\Cref{fig:occupancy} shows the predicted and observed occupancy rates for the different policies, considering both unbiased and biased predictors.
When considering a consistently biased predictor ($\sigma = 0$), both the observed and predicted occupancy rates are identical across all policies in all overestimation scenarios ($\mu \geq 0$).
This matches the trends observed in \Cref{fig:canc} and \ref{fig:aff}, where no cancellations or adjustments occur when $\sigma=0$ and $\mu \geq 0$, hence leading to perfect policy equivalence.
In contrast, in the scenarios with consistent underestimation, policy P performs worse than all others, in terms of both predicted and observed values.
Looking at all instances with $\sigma > 0$, P consistently produces the lowest occupancy rates across all scenarios.
As $\sigma$ increases, the effects of overestimation become less pronounced.
Observed occupancy rates for $\mu=+3$ increase with the standard deviation, while for $\mu=-3$, the occupancy rate only slightly decreases with increasing $\sigma$.
Finally, the predicted occupancy rates remain relatively stable across all scenarios, which is an expected behavior from (near-)optimal schedules.

\section{Conclusions}
\label{sec:losconclusions}


This paper proposes a methodology for simulating the results obtained using a predict-then-optimize approach that employs an \ac{ml} model for parameter prediction.
The main advantage of our methodology is that it does not require collecting data or training \ac{ml} models.
By modeling error metrics using known probability functions, our approach is flexible enough to reproduce a variety of possible error distributions that may occur when training an \ac{ml} model on a real-life application.
Ultimately, our methodology allows decision-makers to assess in advance and at a relatively low cost whether implementing an \ac{ml}-driven solution in a decision process is worthwhile, considering the trade-off between model performance and training costs.
Using this methodology, we investigated the performance of four rescheduling policies used to repair infeasibilities that occur due to poor parameter estimation, as a function of the prediction error.
In particular, we analyzed how errors in patient \ac{los} estimation affect the performance of four rescheduling policies that repair patient admission and surgery schedules.


Even under the assumption of perfect estimation of \ac{los} values, this work highlighted many of the trade-offs that must be considered when planning for elective surgery scheduling: system efficiency versus patient (dis)comfort and operational issues for the hospital.
The different rescheduling policies impact stakeholders in different ways: re-scheduling admissions primarily affects patients and is less impacting for the hospital; transfers are not comfortable for patients, who anyhow manage to get timely care, and are cumbersome activities for both sending and receiving wards; admissions to wards different than ones assigned originally are the least impacting from the patients' perspective but may impact personnel productivity.
From a clinical practice perspective, being affected in any way is preferable to being canceled, as it is also weighted in each objective function of each rescheduling policy.
Overall, we argue that the policy that provides more flexibility (policy C) is generally preferable.
The policy may involve costs not accounted for in the study but provides hospital managers with the best ability to handle schedule changes.
Moreover, results showed that policies CW and T achieve similar results from the hospital's perspective, but the latter achieves so by affecting fewer patients, and this aspect should be considered when evaluating them.
In summary, the more a policy can imply greater hidden costs for the hospital (in descending order: C, T, CW, and P), the more it is beneficial for the system.

Our computational study demonstrated that, from the patients' perspective, overestimating \acp{los} may be beneficial, because it reduces the chance of being canceled \citep{hans2008}.
In some sense, overestimating resembles the rationale behind robust approaches to the problem, as doing so introduces a certain degree of conservatism in the proposed solution, reducing the number of adjustments required to maintain a feasible schedule.
However, when \acp{los} are systematically overestimated, the hospital underperforms in terms of resource occupation and waiting lists expand.
When analyzing results obtained by varying the error $\sigma$ in terms of affected patients, we observed that as $\sigma$ increased, it reduced the negative effect of underestimation, whereas the opposite occurred with overestimation.

To highlight the broader impact of our results, we identified three potential research directions.
First, conducting a cost-benefit analysis using real-world data on operational costs associated with various rescheduling policies would provide a more comprehensive assessment of their feasibility, ultimately helping hospital managers make informed decisions.
Second, extending our analysis to other domains where predict-then-optimize approaches are applicable, such as supply chain management or financial forecasting, would be valuable.
This would require generalizing and adapting our methodology to accommodate different error distributions and decision-making policies.
Finally, integrating our methodology with robust optimization techniques could reveal how such methods interact with estimation errors.
By combining our approach with methods that explicitly account for uncertainty, we could develop decision-making tools that are better suited to handling real-world variability.
Each of these directions would not only strengthen the practical significance of our work but also pave the way for more effective implementation of \ac{ml}-driven solutions in complex decision-making processes involving uncertainty.


\bibliographystyle{cas-model2-names}
\bibliography{bib}

\end{document}